\documentclass[11pt]{article}

\usepackage[preprint]{acl}

\usepackage{times}
\usepackage{latexsym}

\usepackage{fontspec}
\newfontface{\bn}{fontkalpurush.ttf}[
    Script   = Bengali,
    Language = Bengali
]

\newcommand{\bangla}[1]{{\bn #1}}

\usepackage{microtype}
\usepackage{graphicx}

\usepackage{fontawesome5}
\usepackage{booktabs}
\usepackage{alltt}
\usepackage[table]{xcolor}
\usepackage{multirow}
\usepackage{amsmath}
\usepackage{amssymb}
\usepackage{tabularx}

\usepackage{tikz}
\usetikzlibrary{
    arrows.meta,
    positioning,
    fit,
    backgrounds,
    calc
}

\definecolor{cLaya}{HTML}{27EBF5}
\definecolor{cLod}{HTML}{F127F5}
\definecolor{cJev}{HTML}{F5C827}

\definecolor{fLaya}{HTML}{BEF9FC}
\definecolor{fLod}{HTML}{FCCFFD}
\definecolor{fJev}{HTML}{FCECB3}

\definecolor{fGrey}{HTML}{CCCCCC}
\definecolor{fGreyD}{HTML}{4D4D4D}

\newcommand{\laya}[1]{\cellcolor{fLaya}#1}
\newcommand{\lod}[1]{\cellcolor{fLod}#1}
\newcommand{\jev}[1]{\cellcolor{fJev}#1}
\newcommand{\neu}[1]{\cellcolor{fGrey!55}#1}

\newcommand{\layah}[1]{\cellcolor{cLaya!55}\textbf{#1}}
\newcommand{\lodh}[1]{\cellcolor{cLod!42}\textbf{#1}}

\newcommand{\sw}[1]{%
    \textcolor{#1}{\rule[-0.03em]{0.70em}{0.70em}}\,%
}

\newcommand{\swL}{\sw{fLaya}}
\newcommand{\swD}{\sw{fLod}}
\newcommand{\swJ}{\sw{fJev}}
\newcommand{\swG}{\sw{fGrey}}
\newcommand{\swK}{\sw{fGreyD}}

\newcommand{\Drand}{\textsc{d-rand}}
\newcommand{\Dsurf}{\textsc{d-surf}}
\newcommand{\Dsem}{\textsc{d-sem}}
\newcommand{\Dlit}{\textsc{d-lit}}
\newcommand{\Dhard}{\textsc{d-hard}}
\newcommand{\LAR}{\textsc{lar}}

\newcommand{\hlL}[1]{{\setlength{\fboxsep}{1.6pt}\colorbox{fLaya}{#1}}}
\newcommand{\hlG}[1]{{\setlength{\fboxsep}{1.6pt}\colorbox{fGrey}{#1}}}

\newcommand{\pL}[1]{%
    \cellcolor{fLaya}\textbf{#1}%
}

\newcommand{\pJ}[1]{%
    \cellcolor{fJev}\textbf{#1}%
}

\newcommand{\blfootnote}[1]{%
    \begingroup
    \renewcommand\thefootnote{}\footnote{#1}%
    \addtocounter{footnote}{-1}%
    \endgroup
}

\title{
    Literal Lock-In: How Test Language Shifts Literal Attraction\\
    in Typed Decision Models on Bengali Idioms
}

\author{
    Adib Sakhawat\textsuperscript{$*$}
    \quad
    Syed Rifat Raiyan\textsuperscript{$*$}
    \\
    Systems and Software Lab (SSL)
    \\
    Department of Computer Science and Engineering
    \\
    Islamic University of Technology, Dhaka, Bangladesh
    \\
    \texttt{\{adibsakhawat, rifatraiyan\}@iut-dhaka.edu}
    \\[3pt]
    \href{https://github.com/sakhadib/literal-lock-in}{\faGithub{}~\texttt{sakhadib/literal-lock-in}}
    \quad
    \href{https://sakhadib.github.io/literal-lock-in}{\faGlobeAmericas{}~\texttt{sakhadib.github.io/literal-lock-in}}
}

\begin{document}
\maketitle

\blfootnote{\textsuperscript{$*$}Equal contribution.}

\begin{abstract}
Decision models, non-generative scorers that report a probability over
caller-supplied options and emit no text, are increasingly deployed as routers,
classifiers and guardrails. Because they generate nothing, rationales and
chain-of-thought do not apply. We audit three: an encoder scorer, a decoder
scorer and a closed API, entirely through interventions on their input, using
expert-annotated Bengali idioms carrying a gold figurative and a gold literal
gloss. Dropping an idiom's own literal gloss into the option list draws every arm
toward it, taking $65\%$ to $96\%$ of each arm's remaining Bengali errors. A model
$\times$ language matrix over one shared item set shows how far this depends on
the test language: every arm is $1.9$ to $6.4$ times more captured in Bengali than
in English on the same paired entries, the commercial model's apparent resistance
is a property of its test language, and capability does not order the failure.
Calibration degrades in all three arms, and on the open arms
confidence inverts, so gating on it is worse than gating at random. On a
semantics-free option list, where accuracy is at chance, the best-calibrated arm
still reports mean confidence $0.82$. Code and data are available at
\url{https://github.com/sakhadib/literal-lock-in}.
\end{abstract}


\begin{figure}[t]
\centering
\includegraphics[width=\columnwidth]{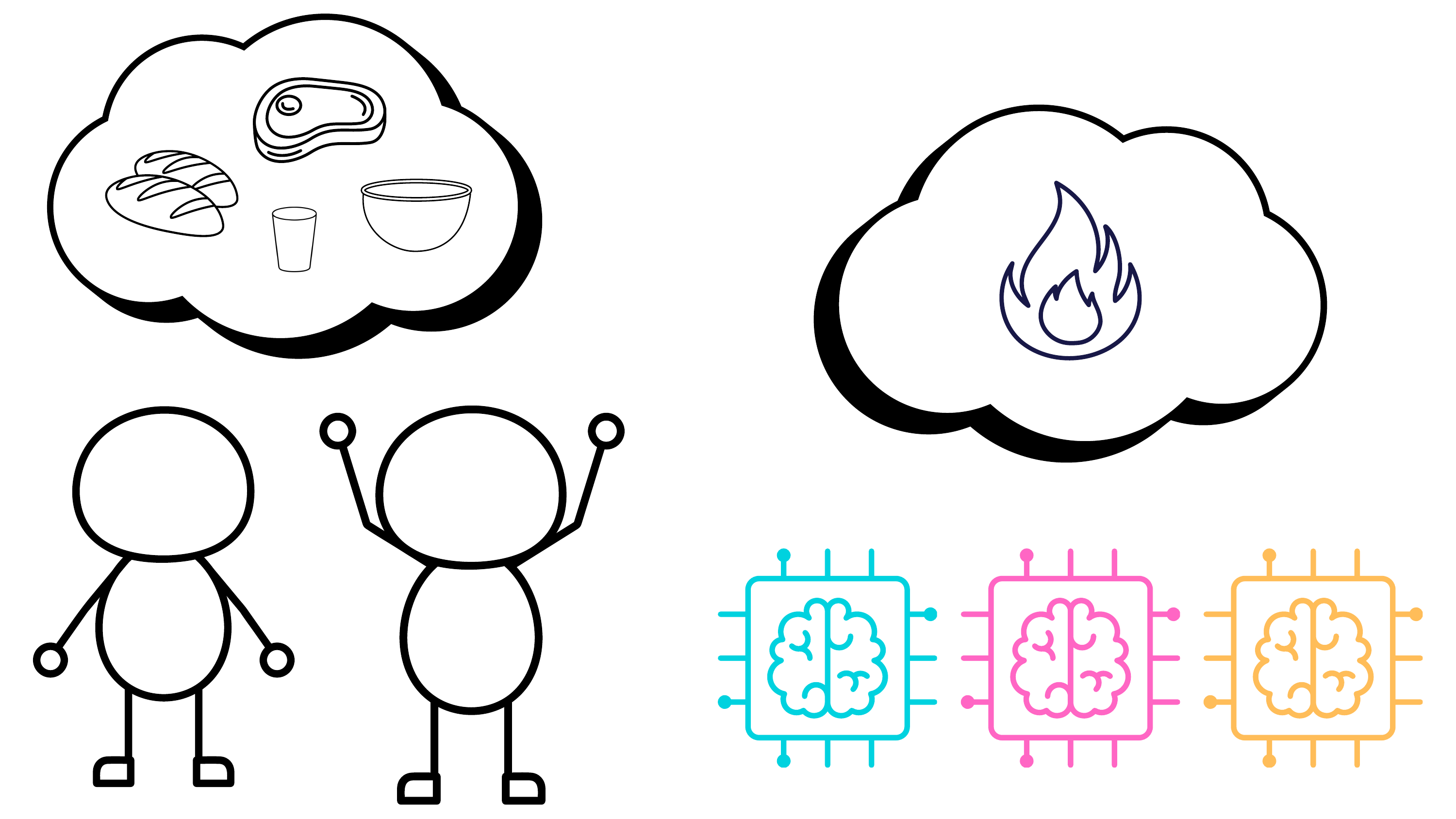}
\caption{\textbf{The model takes the words.}
\swL Laya-ml, \swD Lod-lille and \swJ Jev-1.13 keep these three markers
throughout the paper. The Bengali idiom \bangla{অগ্নিবল} (\emph{agnibal}) is built
from \bangla{অগ্নি} (\emph{fire}) and \bangla{বল} (\emph{strength}) and means
intense hunger. With both readings in the option list, all three models select the
literal gloss in Bengali, at $p=1.00$, $0.97$ and $0.36$. Asked as the paired
English entry, with the English glosses of the same four options, Lod-lille and
Jev-1.13 recover and Laya-ml does not. Across the $3{,}265$ paired items literal
attraction falls from $0.731$, $0.918$ and $0.304$ in Bengali to $0.379$, $0.241$
and $0.047$ in English (\S\ref{sec:capture}). Every item in that pool carries an
English parenthetical inside its Bengali literal option by construction, this one
included; \S\ref{sec:construction} measures what that cue is worth and shows
capture without it.}
\label{fig:teaser}
\end{figure}

\section{Introduction}

A \emph{decision model} is given a question and a list of candidate options and
returns a probability distribution over exactly those options. It does not sample
and it produces no text. We call the interface \emph{typed} because the caller
declares the answer type and the option set in the request, and the response is
guaranteed to be a distribution over exactly those options: the output cannot be
malformed, only wrong. This class has
grown quickly as infrastructure for routing, intent classification, content
gating and tool selection, precisely because it is cheap, bounded and returns a
number a caller can threshold on. An ecosystem analysis of 2{,}170 public
projects finds it already embedded as a reusable decision component across
domains \citep{ling2026wild}, and proposals now place these models at the
adjudication points of autonomous security harnesses
\citep{santos2026pentest}. The threshold is load-bearing.

That bounded interface is also an explainability problem. There is no rationale
to inspect, no chain of thought to measure for faithfulness
\citep{jacovi2020faithfulness,turpin2023unfaithful}, and in the commercial case
no weights at all. What is left is the behavioral surface: the caller controls
the question and the option list, and sees a probability vector. Any audit of
this class has to be built out of that.

The obvious fallback, attributing the decision to input tokens
\citep{ribeiro2016lime,lundberg2017shap}, is both unreliable here
\citep{jain2019attention,adebayo2018sanity,hooker2019roar} and hard to read. How
far an idiom's parts map onto components of its figurative meaning varies by
idiom and is not annotated in the corpus we use
\citep{nunberg1994idioms,gibbs1989kick};
for an expression at the non-decomposable end, a Shapley value over its words is
computable but has no evident linguistic referent. We would rather intervene on
the expression as a unit than report shares we cannot interpret
(Appendix~\ref{app:why}).

We take the option set as the main intervention surface instead. Bengali
\emph{b\=agdh\=ar\=a} (idioms) are well suited to this because the expert
resource we use ships, for each idiom, both a gold figurative meaning and a gold
\emph{literal} gloss: a ready-made, human-certified counterfactual distractor
that no annotation effort of ours produced. Dropping that gloss into the option
list turns a benchmark question into a targeted probe: does the model read the
expression as an idiom, or as its words? Figure~\ref{fig:teaser} shows one item
answered both ways.

We contribute (i) a meaning-choice instrument for decision models in which the \emph{distractor family} is the independent variable and the item, instruction and $k$ are held fixed, requiring no new annotation (\S\ref{sec:instrument}); (ii) the Literal Attraction Rate, measured as a full model $\times$ language matrix, which shows how far capture depends on the test language and that capability does not order it (\S\ref{sec:capture}); (iii) evidence that calibration degrades in all three arms, and that on the two open arms confidence becomes anti-diagnostic, with confidence-gated selective prediction worse than random gating (\S\ref{sec:confidence}); (iv) a large dissociation in option-order dependence between the two scoring designs, with a methodological consequence for anyone evaluating this class (\S\ref{sec:order}); and (v) a semantics-free control, run on all three arms and needing no annotation of any kind, bounding what each model reports when the options carry no information at all (\S\ref{sec:nosignal}).

\section{Related Work}

\paragraph{The decision-model class.} The evaluation literature is still
forming, and reports quality somewhat below large models at far lower cost, with no independent
accuracy advantage from the typed readout itself
\citep{tang2026audit,rao2026rubric,ibrahim2026evaluating,guo2026justask,yu2026visualjev}.
Two strands bear on ours. \citet{ibrahim2026evaluating} report decision-model
\emph{calibration exceeding most LLM baselines}, and \citet{li2026jevjudge}
build the confidence-gated cascade that follows from it; both depend on the
returned confidence being diagnostic, and
\S\ref{sec:confidence}--\S\ref{sec:nosignal} give conditions under which it
is not. Concurrently \citet{sun2026typesafe} find the decision follows an
option's \emph{name} rather than its bound definition, and
\citet{xu2026jevout} flip 61.4\% of correct decisions with short, fluent context
additions chosen to redirect them.
Appendix~\ref{app:positioning} sets out how each relates to our design.

\paragraph{Idioms in NLP.} Idiom resources and tasks are well established
\citep{haagsma2020magpie,tayyarmadabushi2022semeval,chakrabarty2022flute},
including for Bengali \citep{sakhawat-etal-2026-words}, and the difficulty of
figurative language for distributional models is long-noted
\citep{shwartz2019pain}. Psycholinguistics frames the phenomenon as competition
between a literal and a figurative route whose resolution depends on
decomposability and context
\citep{cacciari1988comprehension,gibbs1989kick,titone1999compositional}, and
\citet{oh2026tug} give mechanistic evidence that the competition is real inside
a transformer: causal tracing recovers two pathways carrying the figurative and
literal readings simultaneously, with early layers promoting the figurative one
and a direct route keeping the literal one available.

Our \Dlit{} intervention is the behavioral counterpart of that picture rather
than a replication of it: they can ask \emph{how} the literal route stays
available inside a model they can open, we can only ask \emph{how often} it
wins in one that emits no text and, in one arm, exposes no weights
(Appendix~\ref{app:positioning}).

\paragraph{Evaluating option scorers.} Option-order sensitivity and selection
bias are documented for prompted generative LMs
\citep{pezeshkpour2024order,wei2024selection,zheng2024robust,zhao2021calibrate}.
\citet{sun2026typesafe} relay an independent evaluation harness reporting
option-order sensitivity in Laya; that report is relayed rather than theirs, and
we could not determine from it which checkpoint it covers, so we do not read it
as establishing the behavior of the multilingual checkpoint we audit. What \S\ref{sec:order} adds is the
comparison of option-marker against independent-option scoring over the full
permutation set, and the caveat that permuting a list also relabels it, which
that literature shows is itself a source of bias \citep{zheng2024robust}.

\paragraph{Calibration and selective prediction.} Proper scoring and calibration
measurement are standard
\citep{brier1950verification,gneiting2007proper,guo2017calibration,nixon2019measuring,kumar2019verified},
as is the observation that calibration degrades under distribution shift
\citep{ovadia2019trust,jiang2021know}. Selective prediction gives the operational
test \citep{geifman2017selective,geifman2019selectivenet,traub2024selective}. We
report the combination that matters for deployment: accuracy \emph{decreasing}
with confidence, and a confidence gate underperforming a random one.

\paragraph{Behavioral testing.} Our design follows behavioral checklists,
contrast sets and counterfactual minimal pairs
\citep{ribeiro2020checklist,gardner2020contrast,kaushik2020difference,naik2018stress,jia2017adversarial};
the contribution is carrying that tradition into a model class with no text
output, where it is one of the few instruments left.

\paragraph{Low-resource evaluation.} Bengali has strong encoders and corpora
\citep{bhattacharjee2022banglabert,kakwani2020indicnlpsuite}, and cross-lingual
benchmarks establish the transfer gap
\citep{hu2020xtreme,ruder2021xtremer,adelani2021masakhaner};
\S\ref{sec:crosslingual} localizes our failure to the Bengali side rather than to
figurativeness.

\section{Dataset}
\label{sec:dataset}

\paragraph{Source corpus.} We use \emph{Bangla Bagdhara}
\citep{sakhawat-etal-2026-words}, a corpus of 10{,}361 Bengali idioms
annotated under a 19-field schema by expert consensus.\footnote{The published
count is 10{,}361 and the distributed record file holds 10{,}359; the two missing
rows are lost in the corpus's own export rather than in our loader, which parses
all 10{,}359 without error, so we cannot recover them. Of the 10{,}359, 8{,}876
carry every field the instrument requires.} Each row carries the idiom,
a gold figurative meaning in Bengali, an expert literal gloss, semantic tags,
alternative surface forms certified as meaning-preserving, and a gold carrier
sentence. We add no annotation; every intervention below is assembled from
fields that already exist.

Its own benchmark on the same underlying task reports no model among 30
multilingual and instruction-tuned LLMs above 50\% accuracy, against 83.4\% for
human annotators. Option construction differs, so those figures are not directly
comparable to ours, but they locate the difficulty in the material rather than
in our instrument.

\section{Methodology}
\label{sec:instrument}

\begin{figure*}[t]
\centering
\includegraphics[width=\textwidth]{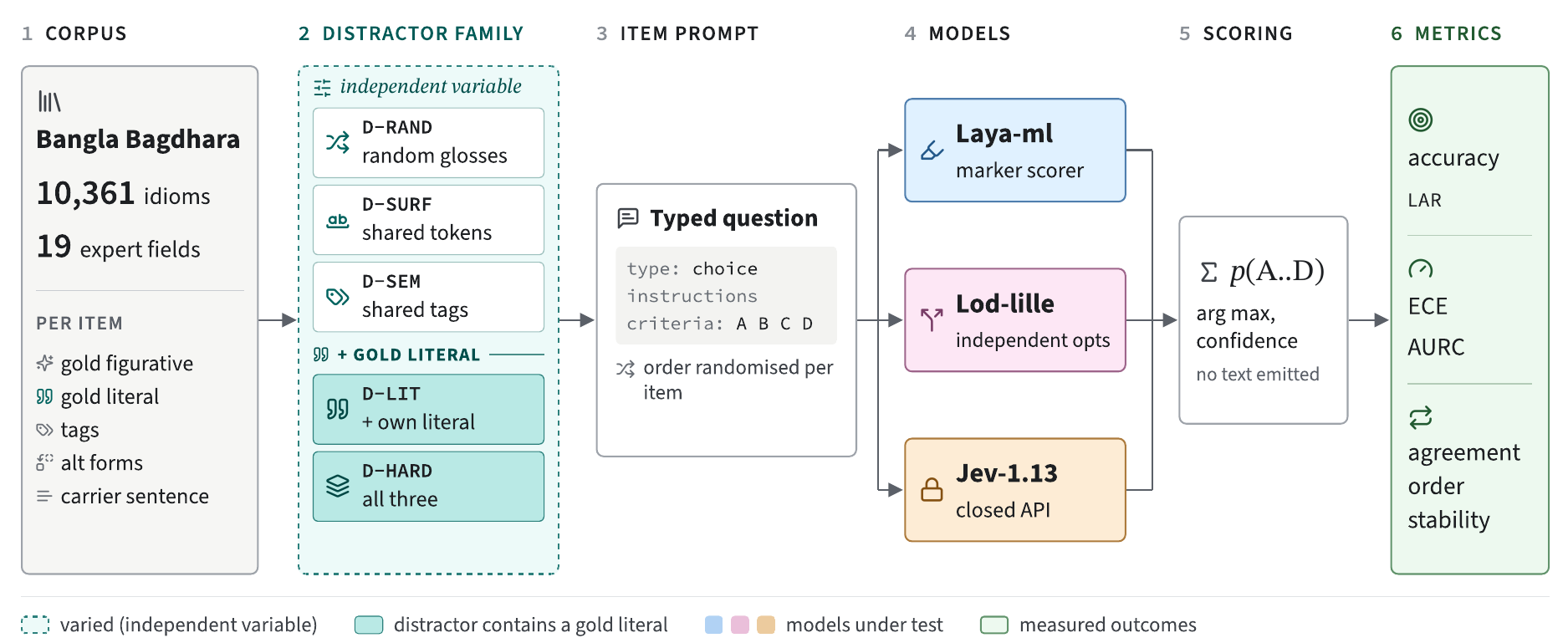}
\caption{\textbf{The instrument in one view.}
The audit proceeds from the \textbf{Bangla Bagdhara} corpus through six
stages: \textbf{(1) corpus}, which supplies 10,361 expert-annotated idioms
with gold figurative and literal glosses; \textbf{(2) distractor family},
the variable E1 manipulates, with $\textsc{d-lit}$ and $\textsc{d-hard}$
injecting the expert literal gloss; \textbf{(3) item prompt}, which fixes
the typed question, $k=4$ options, instructions, and per-item order seed;
\textbf{(4) models}, comprising Laya-ml, Lod-lille, and Jev-1.13;
\textbf{(5) scoring}, which records the four-option probability vector,
$\arg\max$, and confidence without requiring generated text; and
\textbf{(6) metrics}, which evaluate accuracy, LAR, ECE, AURC, agreement,
and order stability. The figure's visual encoding distinguishes the
varied distractor family, literal-bearing conditions, models under test,
and measured outcomes.}
\vspace{-2mm}
\label{fig:pipeline}
\end{figure*}

Figure~\ref{fig:pipeline} sets out the instrument end to end: what the corpus
supplies, what each condition changes, and what comes back.

\subsection{The meaning-choice task}

Given an idiom, choose its gold figurative meaning from $k=4$
candidates. Chance is $0.250$ throughout. The \textbf{distractor family is the
independent variable}; the item, the instruction, $k$, and per-item option-order
randomization (seeded by item id and experiment) are held fixed across families,
so a paired test across families compares one item under different option lists.

\subsection{Distractor families and measures}

The five families are \Drand{} (other idioms' glosses at random), \Dsurf{}
(glosses sharing surface tokens with the gold), \Dsem{} (glosses from idioms
sharing semantic tags), \Dlit{} (one distractor replaced by the idiom's
\emph{own} expert literal gloss) and \Dhard{} (all three pressures at once).
Appendix~\ref{app:experiments} gives each sampling rule, and
Table~\ref{tab:bnitem} one complete item in the original script.

For the families containing a literal gloss, with $\ell_i$ the literal option
and $g_i$ the gold, the Literal Attraction Rate is
\[
\LAR = \frac{1}{N}\sum_{i=1}^{N} \mathbf{1}\!\left[\arg\max_o p_i(o) = \ell_i\right].
\]
\LAR{} is not $1-\text{accuracy}$: it isolates the errors that are specifically
a literal reading, and has no floor, since a model that never reads literally
scores $0$ while still being wrong.

\subsection{Interventions and statistics}

Seven interventions vary one thing each: the distractor family (E1), the
idiom's surface form (E2), the idiom as a unit (E3), the task language (E4),
nothing (E5, derived from E1), the option text (E6) and the option order (E7).
Appendices~\ref{app:experiments} and \ref{app:payload} give what each holds
fixed and the typed question they share.

All intervals are 95\% percentile bootstrap over items (2{,}000 resamples);
paired comparisons use McNemar's exact test; ECE uses 15 equal-width bins and
AURC is reported against a random-gate baseline. The two open arms were built
independently from a written contract, and on all 4{,}440 overlapping rows the
gold key, gold position, idiom string and literal-distractor key agree at
$1.000$, so every paired test is on identical questions
(Appendix~\ref{app:repro}).

\subsection{Models}

Three models span the design space (Table~\ref{tab:models}); none has an
accompanying paper, so we identify them by their released
artifacts.\footnote{\textbf{Laya-ml}: \texttt{convaiinnovations/laya}, subfolder
\texttt{multilingual}, on the Hugging Face Hub; the same repository ships the
English-only checkpoint used as the control in \S\ref{sec:crosslingual}.
\textbf{Lod-lille}: \texttt{mrn-dk/lod-lille-0.6B}, same hub.
\textbf{Jev-1.13}: \texttt{typesafe/jev-1.13}, served over OpenRouter's
\texttt{/api/alpha/decisions} endpoint, with no weights, training data or
architecture published.}
\textbf{Laya-ml} is a bidirectional encoder scoring all options together from
marker positions. \textbf{Lod-lille} is a decoder with independent-option
scoring, each option attending only to the shared state, the question and itself
from the same start position, which its model card reports as order-independent
by construction in \texttt{fp32}; \S\ref{sec:order} finds that reordering and relabeling the list together still
move its scores.
\textbf{Jev-1.13} is a closed commercial model behind a decisions API whose
documentation names English as its primary and best-supported language; we know
its interface and nothing else. Appendix~\ref{app:payload} gives the call each one
receives.

\section{Analysis}
\label{sec:analysis}

\subsection{Literal capture, and what it tracks}
\label{sec:capture}

\begin{figure*}[t]
\centering
\includegraphics[width=0.72\linewidth]{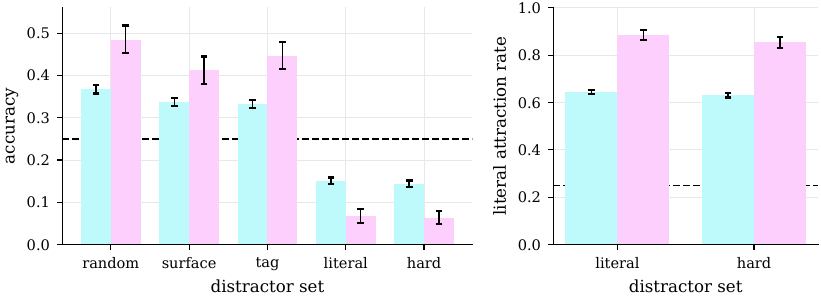}
\caption{The distractor ladder (left) and literal attraction rate (right) for
the two open arms. \swL Laya-ml ($n=8{,}876$), \swD Lod-lille ($n=888$); the
black dashed line marks the four-option chance level ($0.250$) in both panels;
error bars are 95\% bootstrap CIs. Surface- and tag-matched distractors cost a
few points; the idiom's own literal gloss (\Dlit) collapses both models far below
chance, and the \emph{stronger} model harder.}
\label{fig:ladder}
\end{figure*}

Table~\ref{tab:ladder} gives the ladder for the two open arms on the 888 items
they share; Figure~\ref{fig:ladder} plots Laya-ml's full set alongside
(Appendices~\ref{app:ladder}, \ref{app:strata}). Lod-lille is the stronger
model on every non-literal condition, by up to $+0.133$. Insert the idiom's own
literal gloss and the ordering \emph{reverses}: Lod-lille falls to $0.066$, half
of Laya-ml's already-below-chance $0.136$, and the same holds for \Dhard.

\begin{table}[t]
\centering\small
\renewcommand{\arraystretch}{1.12}
\begin{tabularx}{\columnwidth}{@{}>{\raggedright\arraybackslash}X>{\centering\arraybackslash}p{1.05cm}>{\centering\arraybackslash}p{1.05cm}>{\centering\arraybackslash}p{1.25cm}}
\toprule
\textbf{Distractors} & \textbf{Laya} & \textbf{Lod} & $\boldsymbol{\Delta}$ \\
\midrule
random & 0.351 & \lodh{0.484} & \neu{$+0.133^{***}$} \\
surface & 0.339 & \lodh{0.412} & \neu{$+0.073^{***}$} \\
tag & 0.325 & \lodh{0.446} & \neu{$+0.120^{***}$} \\
\midrule
$+$ literal & \layah{0.136} & \lod{0.066} & $-0.070^{***}$ \\
hard & \layah{0.149} & \lod{0.063} & $-0.086^{***}$ \\
\midrule
\LAR{} (lit.) & 0.658 & \lodh{0.885} & \\
\LAR{} (hard) & 0.615 & \lodh{0.855} & \\
mean $p$(lit.) & 0.460 & \lodh{0.726} & \\
mean $p$(gold) & \layah{0.193} & 0.127 & \\
\bottomrule
\end{tabularx}
\caption{\textbf{The distractor ladder and literal attraction.}
Every cell is the 888 items the two arms share, so the printed values and
$\Delta$ have one base ($\Delta$ is computed before rounding); Laya-ml's full-set figures are in
Appendix~\ref{app:ladder} and differ by at most $0.016$
(Appendix~\ref{app:subset}). McNemar is paired on those items
($^{***}$ $p<0.001$). \swL Laya-ml, \swD Lod-lille; the darker fill marks the
higher value in a row. The sign of $\Delta$ reverses at the literal boundary: the
stronger arm leads on the first three families and trails on the two containing a
literal gloss.}
\vspace{-2mm}
\label{tab:ladder}
\end{table}

Within this pair, then, \LAR{} measures something accuracy does not: a model
can be selected as ``better'' on a conventional benchmark while being worse at
what that benchmark proxies for. Laya-ml and Lod-lille answer byte-identical questions in the same language, so
the crossover between them is a paired result. The error is not diffuse:
the decomposition under \Dlit{} is gold / literal / other $= 0.136$ / $0.658$ /
$0.206$ for Laya-ml and $0.066$ / $0.885$ / $0.048$ for Lod-lille, both on those
888 items, and on the clearest of them Laya-ml places $p = 1.000$ on the literal
gloss. Laya-ml's full-set decomposition, $0.151$ / $0.644$ / $0.205$, is in
Appendix~\ref{app:ladder}.

\paragraph{Capture tracks the language, not capability.} The arms above ran the
literal condition in Bengali, while the commercial arm's headline $\LAR = 0.047$ is
\emph{English}, so the two are not comparable as they stand. We therefore ran the
full model $\times$ language matrix on the 3{,}265-item shared pool, every cell
built by one function from one seed (Table~\ref{tab:matrix},
Figure~\ref{fig:matrix}); Jev-1.13's English cell lands at $0.0472$
$[0.0401, 0.0545]$, matching its value in the English ladder of
Appendix~\ref{app:ladder}.

Two things follow. First, \textbf{every arm is far more captured in Bengali than in
English on the same paired entries}: $0.731 \rightarrow 0.379$, $0.918 \rightarrow
0.241$ and $0.304 \rightarrow 0.047$, all paired McNemar $p < 10^{-100}$. Second,
\textbf{capture is not monotone in capability}: Lod-lille, the stronger open arm on
random distractors, is the more captured in Bengali, and Jev-1.13 the least. The
open arms fall below chance here; Jev-1.13 does not, but loses $0.390$ of accuracy
against its own English run. The pool is not neutral and its Bengali literal option
carries a cue the English one lacks; on ladder items without it,
Bengali capture is still $0.581$ and $0.871$ (Appendices~\ref{app:matrix},
\ref{app:construction}).

\paragraph{The capability ordering.} One scored condition is matched across all three
arms, Bengali idiom with random distractors, and on it the commercial model is far
ahead, $0.736$ against $0.367$ and $0.484$ (Appendix~\ref{app:threeway}). That is
the ordering the calibration results below should be read against, and the one
capture does not follow.

\subsection{Confidence is anti-diagnostic}
\label{sec:confidence}

Table~\ref{tab:conf} gives the mean probability on the chosen option, split by
outcome, on the literal-bearing families. Both open models are \emph{more
confident when captured than when correct}, Lod-lille extremely so ($0.787$
against $0.453$), and their reliability curves slope \emph{downward} under \Dlit{}
(Figure~\ref{fig:reliability}, Appendix~\ref{app:selective}).

\begin{table}[t]
\centering\small
\setlength{\tabcolsep}{4pt}
\begin{tabular}{lccc}
\toprule
\textbf{Outcome} & \textbf{Laya} & \textbf{Lod} & \textbf{Jev} \\
\midrule
correct          & 0.432 & 0.453 & \textbf{0.935} \\
captured         & \textbf{0.596} & \textbf{0.787} & 0.737 \\
other error      & 0.421 & 0.399 & 0.746 \\
\midrule
ECE (\Drand)     & 0.071 & 0.073 & 0.005 \\
ECE (\Dlit)      & 0.385 & 0.679 & 0.014 \\
AURC (conf.)     & 0.924 & 0.989 & 0.013 \\
AURC (random)    & 0.849 & 0.934 & 0.069 \\
\bottomrule
\end{tabular}
\caption{Confidence by outcome, pooled over the two literal-bearing families
\Dlit{} and \Dhard, with calibration below.
Bold marks, per model, whether it is more confident when correct or when captured.
Each column is on that arm's own items, so compare pattern rather than level.
Both open arms are more confident when captured, and their confidence-gated AURC
is \emph{worse} than the random-gate baseline on the row below it, which inverts
selective prediction. Both AURC rows are the \Dlit{} condition, and the random-gate
value is the expected AURC of a confidence signal drawn at random on the same
items.}
\label{tab:conf}
\end{table}

ECE depends on the binning \citep{nixon2019measuring}, so we also report the
binning-free multiclass Brier score: $1.030$ and $1.391$ for the open arms under
\Dlit{} against $0.750$ for a uniform four-way distribution, so their reported
probabilities are worse than declining to choose. Equal-mass bins move no ECE by
more than $0.011$ (Appendix~\ref{app:selective}).

Selective prediction therefore inverts on both open arms: restricting Laya-ml's
\Dlit{} to its most confident 5\% gives $0.007$ accuracy against $0.151$ at full
coverage, and Lod-lille answers every item in its most confident 35\% incorrectly
(Appendix~\ref{app:selective}). It is not an artifact of one confidence signal:
Lod-lille's \emph{separate confidence head}, which estimates whether the answer is
in the option list at all, shows the same inversion ($0.631$ against $0.295$ over
both literal-bearing families), and
it survives AUGRC
\citep{traub2024selective}, which matters here because AURC aggregates risk
conditioned on the accepted set and so over-weights failures made at high
confidence, which is exactly where our inversion lives. For a random confidence
signal AUGRC is half the error rate, so the baseline needs no simulation: under
\Dlit{} the open arms score $0.448$ and $0.491$ against $0.425$ and $0.467$, worse
than random on that measure as on AURC.

Calibration degrades in all three arms; only the inversion is two-arm. Jev-1.13
looks well calibrated in English (ECE $0.014$, Brier $0.107$) and is not on the
same items in Bengali (ECE $0.135$, Brier $0.604$), where its advantage over a
random gate falls from $5.1\times$ to $1.4\times$. It is not inverted, though,
staying more confident when correct ($0.724$) than when captured ($0.644$), so the
sign depends on the arm. This does not contradict \citet{ibrahim2026evaluating}:
in-distribution calibration need not survive a targeted distractor shift, and it
fails in the costly direction, since a confidence-gated cascade
\citep{li2026jevjudge} accepts without review exactly the tail where the open arms
are \emph{least} accurate
(Appendix~\ref{app:positioning}).

\subsection{Semantics-free control, on all three arms}
\label{sec:nosignal}

The \textsc{o4} ablation replaces every option's text with a bare ordinal label,
\bangla{অর্থ}~\texttt{1}--\texttt{4} for the Bengali arms and
\texttt{meaning}~\texttt{1}--\texttt{4} for the commercial one, so the four options are distinguishable but carry no meaning. With the gold
position randomized and no option carrying information, expected accuracy is
$0.25$ and whatever confidence survives above it is overconfidence, with no modeling
assumption needed to read it. It needs no gloss and no literal annotation, so it
is the one condition we can put to all three arms without the pool restriction;
the commercial arm's run is in English, so the labels differ in language from the
other two while carrying the same absence of meaning
(Figure~\ref{fig:nosignal}).

The ordering reverses against Table~\ref{tab:conf}. The best-calibrated model
under real options is the worst when the options carry nothing: Jev-1.13 reports
$0.815$ mean confidence at $0.254$ accuracy, an overconfidence of $0.561$, which
we give as a gap rather than a binned ECE because the gap is a lower bound on ECE
and needs no binning choice to defend. Lod-lille is the only one whose confidence
falls close to the $0.25$ a calibrated model would report ($0.282$).

Jev-1.13 exposes a second field meant for exactly this case, a confidence that the
answer is present in the option list at all, and it does not rescue the reading:
on \textsc{o4} it reports $0.754$ against $0.946$ with full glosses. The reference
is not the $0.25$ chance level of the top probability, since this field reports
presence, so on a list where the answer provably is not present an informed value
is near zero. The pattern matches \citet{zhong2026rejection} on a different task,
with a scope condition that cuts both ways and that
Appendix~\ref{app:positioning} sets out: a \emph{verification}-typed call recovers
what the choice type does not, and we have not tested whether that holds here.
What the control shows is what a caller of this answer type is told, which is why
we read it as the most portable finding: it needs no annotation and no matched
language. McNemar between the two
open arms under \textsc{o4} is not significant ($p = 0.76$), so the instrument's
null behaves alike in both.

\subsection{Option order is architecture-specific}
\label{sec:order}

We run every item under all $4! = 24$ orderings of a fixed option set and record
the chosen gloss's identity, recovered as
$\mathrm{perm}[\mathrm{slot}(\hat{y})]$; Appendix~\ref{app:position} explains why
the slot letter is the wrong unit for the \emph{answer}.

One caveat applies to everything here. Permuting
the list does two things at once: it moves each gloss and it changes which letter
names it. They cannot be separated in this design, and the literature gives reason
to think the second matters, since \citet{sun2026typesafe} rebind option names to
fixed definitions and find the decision follows the name, while
\citet{zheng2024robust} trace most selection bias in multiple choice to token-level
priors on the option IDs. What we measure is therefore sensitivity to display order
and labeling \emph{together}, which is what a caller varies when they reorder a
list, not a clean effect of position. Separating them needs a condition that moves
each gloss while it keeps its own letter, which we did not run.

Table~\ref{tab:order} gives six measures of that sensitivity, each on an item
sample drawn separately per arm. The marker-based
scorer changes its chosen gloss on $83.7\%$ of items against $10.7\%$ for the
independent-option scorer and $13.3\%$ for the commercial one, a $7.8\times$
difference in the rate of answer changes and a $12\times$ difference in the
within-item range of $p$(gold). Each scoring design is represented by exactly one model, so the
design cannot be separated from the backbone or the capability of the model
carrying it; we read the pattern as consistent with the scoring architecture
rather than as demonstrated to follow from it.

\paragraph{The invariance claim does not survive as stated.} Lod-lille is
documented as order-independent by construction, each option being scored from the
same start position so that no option sees another. Mapping every slot's
probability back to the option it displayed, a single option's own score travels
$4.0 \times 10^{-2}$ at the median item and never stays below $10^{-4}$, orders of
magnitude above \texttt{fp32} noise (Appendix~\ref{app:position}). Given the
caveat above we state the conclusion narrowly: a caller who permutes an option
list does not get identical probabilities back. Whether the residue is positional or comes from the letters
being reassigned we cannot say, and the second would be no less a failure of the
guarantee than the first.

A slot preference follows, measured over all five families of the full ladder,
where the gold position is randomized: Laya-ml takes slot B $37.2\%$ of the time
against $14.6\%$ for slot A (Appendix~\ref{app:position}). Here too slot and letter
covary, so it is as consistent with a prior on the token \texttt{B} as with a
preference for the second position. Either way, an evaluation of this class that
fixes option order is partly measuring something other than the options. All
results here randomize order per item, seeded by item id and experiment, and we would treat that
as a default rather than a robustness check.

\subsection{The option side is read lexically, and the idiom is not held as a unit}
\label{sec:bagofwords}

Two interventions probe how each side of the item is read, and both carry scope
conditions we state with the result.

On the option side, scrambling the words inside every option costs almost nothing:
writing \emph{signal} for accuracy above chance as a fraction of the full-gloss
condition's, the three arms retain $0.928$, $0.944$ and $1.005$ of it. The
companion truncation is weaker than it looks, since the median Bengali gloss is
four words and truncation to six is a no-op on $77.8\%$ of options. The scramble
is the result, and it converges with \citet{sun2026typesafe} from the other
direction; the scope condition is that the series runs on random distractors,
where recognizing the right words already suffices, so it shows composition is not
\emph{required} here rather than that these models cannot compose
(Appendix~\ref{app:options}).

On the idiom side, E3 perturbs the expression while holding the option list fixed.
It covers only idioms of three words or more, so every manipulation changes the
string on $99.7\%$ of items or better. Accuracy cannot carry the argument, because
the condition is \Dlit{} and the intact accuracies on E3's own items, $0.157$ and
$0.072$, already sit below chance. Agreement with the intact answer has no such
floor, and that is where the claim rests: under word-shuffling the open arms
return the same option on $0.742$ and $0.917$ of items, at least as often as under
meaning-preserving rewrites ($0.657$, $0.677$; Appendix~\ref{app:surface}). The one manipulation that moves both is the gold
carrier sentence, $+0.080$ and $+0.117$, so context rather than the idiom string
carries the usable signal. That cuts both ways, since \citet{xu2026jevout} flip
61.4\% of correct decisions with short fluent additions chosen to redirect them
(Appendix~\ref{app:unit}).

\subsection{What the item construction could explain}
\label{sec:construction}

The literal gloss is an annotated field rather than a string we wrote, and in this
corpus it is often lexically close to the idiom. Three properties could let a
lexical matcher select it without reading anything. The idiom is a single word on
$42.3\%$ of items, and on $18.0\%$ the literal gloss contains the idiom verbatim.
In the matrix pool it is the only option written partly in Latin script, since no
Bengali figurative gloss in the corpus contains a Latin character. And it is longer
than the gold gloss on $67\%$ of items, though longest of all four on only
$40.2\%$.

Two of the three buy measurable capture (Table~\ref{tab:construction}): verbatim
overlap on both arms, and the Latin-script cue on Laya-ml, with Lod-lille's
interval covering zero. Length is the exception and needs separate handling, since
the subset below does not remove it. Splitting on whether the literal option is
the longest of the four moves \LAR{} by $-0.034$ $[-0.055,-0.013]$ on Laya-ml and
$+0.020$ $[-0.024,+0.064]$ on Lod-lille, so Laya-ml is slightly \emph{less}
captured when it is: a real asymmetry in the material, not a route to capture.

Restricting to the $2{,}480$ items that are multiword, whose gloss does not
contain the idiom and carries no parenthetical, removes both lexical cues and
narrows the length gap to $36.6$ against $32.4$ characters. \LAR{} is still
$0.560$ and $0.891$ against a chance level of $0.250$, and the length split stays
null; Appendix~\ref{app:items} has the rest, including \LAR{} by idiom length.

\begin{table}[t]
\centering\small
\renewcommand{\arraystretch}{1.12}
\begin{tabularx}{\columnwidth}{@{}>{\raggedright\arraybackslash}X>{\centering\arraybackslash}p{1.4cm}>{\centering\arraybackslash}p{1.4cm}@{}}
\toprule
& \textbf{Laya} & \textbf{Lod} \\
\midrule
\multicolumn{3}{@{}l}{\emph{Bengali \LAR{} under \Dlit, by cue}} \\
idiom not in gloss      & 0.616 & 0.866 \\
idiom in gloss          & 0.767 & 0.986 \\
\quad difference        & $+0.151$ & $+0.120$ \\
\addlinespace
no Latin-script cue     & 0.581 & 0.871 \\
Latin-script cue        & 0.746 & 0.909 \\
\quad difference        & $+0.165$ & $+0.037$ \\
\addlinespace
\multicolumn{3}{@{}l}{\emph{both lexical cues absent} ($n=2{,}480/266$)} \\
\LAR                    & \textbf{0.560} & \textbf{0.891} \\
accuracy                & 0.187 & 0.079 \\
\bottomrule
\end{tabularx}
\caption{\textbf{Two lexical cues are real; neither carries the result.}
\swL Laya-ml, \swD Lod-lille. Each cue would let a lexical matcher pick the
literal option without reading it. Verbatim overlap moves \LAR{} on both arms and
the Latin-script cue on Laya-ml only, its Lod-lille interval covering zero. The
last block keeps items that are multiword, whose literal gloss does not contain
the idiom and carries no English parenthetical, so neither lexical cue is
available: \LAR{} stays far above the $0.250$ chance level on both arms. Length is
not removed by that subset and is reported separately in \S\ref{sec:construction},
where it does not predict capture. The commercial arm cannot be split this way,
since its only Bengali literal run is the matrix pool, where every item carries an
English parenthetical by construction.}
\label{tab:construction}
\end{table}

\subsection{Failure localizes to Bengali, in every arm}
\label{sec:crosslingual}

E4 asks the same question of an ordinary benchmark item rather than a
literal-bearing one: hold the model fixed, swap the task language, keep the
distractors random. Every arm moves by a large margin, Laya-ml $+0.246$,
Lod-lille $+0.380$ and Jev-1.13 $+0.230$, each paired on that arm's own shared
items (all McNemar $p<0.001$). Lod-lille reaches $0.864$ in English against
$0.484$ in Bengali, so the figurative concepts appear largely available to it and
the Bengali route is where the loss occurs, consistent with its documented
English-only training evaluation. Laya-ml's English-only checkpoint on Bengali
script sits at $0.283$, barely above chance, against $0.371$ for the multilingual
one on the same items, so the scorer does read Bengali where the checkpoint can. Here too the
English condition presents the annotator's nearest English idiom rather than a
translation (Appendices~\ref{app:crossling}, \ref{app:construction}).

\section{Discussion}

Three cautions follow for anyone deploying this class. A returned probability is
worth little without a measurement of what the model returns on a degenerate
option list, a case that arrives from an upstream bug or an empty retrieval rather
than from an adversary. A benchmark score is not by itself evidence against
literal capture, since on our open arms the two move in opposite directions. And
an evaluation that fixes option order is partly reporting position and labeling,
which \S\ref{sec:order} shows cannot be separated here. None of this rebuts the
cascade designs, whose cost and latency advantages are established
\citep{tang2026audit,rao2026rubric}; it bounds them, since the property they rest
on degrades first and silently.

\section{Conclusion}

Decision models are being placed at decision points, and the interface that makes
them attractive is also what makes them hard to audit: no rationale, no chain of
thought, and in the commercial case no weights. We have argued that the caller's
input is an adequate substitute, since every result here comes from changing the
option list or the expression it asks about and reading the probability vector
back, which needs no new annotation and works unchanged on a closed, text-free
model.

One substituted option is enough. The idiom's own expert literal gloss draws all
three arms toward the literal reading, taking $65\%$ to $96\%$ of what each arm
gets wrong in Bengali, and the model $\times$ language matrix separates two
explanations that are easy to conflate: capture falls by $0.26$ to $0.68$ in every
arm when the paired entries are asked in English, so the commercial model's
apparent resistance belongs to its test language rather than to the model.
Capability does not order the failure: the middle arm by accuracy is the most
captured in Bengali. Nor is any of this visible to the caller: confidence inverts on
the open arms under the literal shift, calibration degrades on all three, and on an option
list stripped of semantics the arms still report between $0.28$ and $0.82$.

What the design cannot settle is set out in Limitations, and the cues that could
have produced this result are measured in \S\ref{sec:construction} rather than
assumed away. We release the instrument and the per-item outputs.

\section*{Limitations}

All three models are evaluated zero-shot, and this is the limitation we would
weigh most heavily after the one below. Decision models are trained on a
particular set of decision tasks, and the open arms' own documentation reports
that base checkpoints are weak on tasks outside that set. Some part of what we
measure may therefore be out-of-distribution task transfer rather than literal
capture specifically. The \textsc{o4} control bounds this from one side, since
it shows both open arms at chance accuracy with no detectable difference between
them, and the
\Drand{} to \Dlit{} contrast is within-model and within-item, which removes any
fixed task-transfer offset. Neither argument rules the confound out. A single
lightly fine-tuned arm would settle it, and we have not run one.

Laya-ml's run had its per-bucket temperatures unset, so its ECE figures are raw
and we do not describe the arms as representing distinct calibration philosophies:
both open arms are effectively untempered and comparable on that axis, and
Jev-1.13's internal calibration is unknown to us. Laya-ml also ran with
\texttt{bf16} requested on a Turing card, which has no \texttt{bf16} tensor-core
path, so the arithmetic falls back to \texttt{fp32} CUDA cores
(Appendix~\ref{app:repro}); a reader reproducing on
later hardware is not reproducing the same numerics.

Outside the capture matrix, which it ran on the full $3{,}265$-item pool,
Lod-lille ran a fixed 10\% subset (888 items), so its intervals there are roughly
$\pm 3$ points and its stratified cells are thin. Appendix~\ref{app:subset}
bounds the subset bias at $0.016$ and shows it changes no sign or ordering. E3
on the Lod-lille arm is 180 items and several of its perturbation contrasts are
not individually significant. E7 covers 992 / 299 / 300 items rather than the
full sets, drawn separately per arm, so its cross-arm comparison is unpaired.

Three of the interventions are weaker than their headings suggest, and we would
rather say so than have it found. E7 cannot separate position from labeling,
because permuting a list reassigns the letters with it; the condition that would
separate them moves each gloss while it keeps its own letter, and we did not run
it. E3 runs under \Dlit, where intact accuracy is already below chance, so its
accuracy rows have little room to fall and the conclusion rests on the
intact-answer agreement instead; running it under \Drand{} would let accuracy
carry the argument too. E6 runs under \Drand, where matching the right words is
already enough to answer, so it shows composition is not required rather than
that it is unavailable; the scramble under \Dlit{} or \Dsem{} is the test that
would separate those, and it is unrun. None of the three affects
\S\ref{sec:capture}, \S\ref{sec:confidence} or \S\ref{sec:nosignal}.

The matrix in \S\ref{sec:capture} runs on the 3{,}265 items that carry a usable
English literal gloss, which is not a neutral sample: for Laya-ml, in-pool \LAR{}
is higher than out-of-pool while \Drand{} accuracy is unchanged, so the absolute
rates there are higher than they would be corpus wide. That difference is not
generic pool selection. Pool membership requires an English parenthetical inside
the Bengali literal gloss, and Appendix~\ref{app:construction} shows the
parenthetical is itself worth $+0.165$ of \LAR{} on Laya-ml, which is most of the
in-pool elevation. The language comparison is paired within item and within model,
so the selection affects the levels rather than the direction. The matrix also covers
the literal condition only; we did not run the other distractor families in both
languages. The confidence results in \S\ref{sec:confidence} remain two-arm,
because the commercial arm never shows the inversion, not because it was
unmeasured: \S\ref{sec:confidence} reports its Bengali calibration from the matrix
outputs.

The two language cells are also not symmetric in construction, and this is the
limitation we would flag hardest. They are paired entries rather than identical
items: the Bengali state is the idiom, while the English state is the annotator's
nearest English idiom rather than a translation of it. The two literal glosses are
drawn from one annotated field in different ways, which leaves the Bengali literal
option both lexically closer to its own state and, because it retains its English
parenthetical, the only option in the list written partly in Latin script.
Appendix~\ref{app:construction} sets out all four asymmetries we could find and
tests the two that recorded outputs can reach. Restricted to the $1{,}429$ items
whose Bengali literal gloss shares no content word with the idiom, the capture gap
is $+0.358$, $+0.610$ and $+0.310$ against headline gaps of $+0.352$, $+0.677$ and
$+0.256$. Restricted to the ladder items whose literal gloss carries no English
parenthetical at all, $5{,}485$ for Laya-ml and $560$ for Lod-lille, Bengali capture
is still $0.581$ and $0.871$, against a chance level of $0.250$. Neither cue accounts for the
result, though the second is worth a real $+0.165$ on one arm and we report it as
such. What we cannot do without new runs is make the two cells symmetric by
construction; a translation-paired corpus, or a Bengali cell with the parenthetical
stripped, would settle what we can only bound.

More broadly, this is one corpus, one language, two small open models and one
commercial endpoint. The semantics-free control is the only result here that we
would expect to transfer as stated; the rest should be read as findings about
these systems on this material, and as an instrument other people can point at
theirs.

Distractors are drawn from the same corpus, so difficulty is corpus-relative
rather than absolute. Jev-1.13 is a commercial endpoint and may change without
notice; our results are a snapshot, and we report the exact call count and cost
in Appendix~\ref{app:repro} so the snapshot is at least auditable.

\section*{Ethics Statement}

The corpus is a published, expert-annotated resource of idioms \citep{sakhawat-etal-2026-words}; it contains no
personal data and no content about identifiable individuals, and it is
distributed under CC BY-SA 4.0. Our interventions
rewrite option lists and perturb idiom strings, and produce no new annotation and
no generated text.

The work audits two openly released models and one commercial API. We report
failure modes in all three, including the vendor's, and we have taken care to
separate what is measured from what is inferred. Where a reading of our own data
would have been more flattering than the data supports, we say so in the text:
\S\ref{sec:capture} states the pool selection effect against our cleanest number,
and Appendix~\ref{app:construction} reports a construction cue that accounts for
part of the headline effect on one arm. Commercial endpoints are mutable, so the
Jev-1.13 results should be read as a dated snapshot rather than a standing
property of the product.

Bengali is under-served relative to its speaker population, and we think an audit
that locates a failure in the language rather than in the task is the more useful
kind to publish. The instrument, the per-item outputs and the analysis code are released under
CC BY-SA 4.0, the corpus's own license, and the two open checkpoints are used
under the terms their model cards declare;
Appendix~\ref{app:repro} is the run record they correspond to.

\bibliography{references}

\appendix

\section{The typed question}
\label{app:payload}

Every call in this paper sends one typed \texttt{choice} question and receives
one probability vector. Nothing else is exchanged: there is no prompt template,
no system message, no sampling parameter and no generated text. This is the
whole interface, which is why every intervention in this paper acts on one of the
two things the caller supplies: the state, or the option set.

\begin{table}[t]
\centering\small
\renewcommand{\arraystretch}{1.15}
\begin{tabularx}{\columnwidth}{@{}>{\raggedright\arraybackslash}p{1.35cm}*{3}{>{\centering\arraybackslash}X}}
\toprule
& \textbf{\laya{Laya-ml}} & \textbf{\lod{Lod-lille}} & \textbf{\jev{Jev-1.13}} \\
\midrule
\textbf{backbone} & mmBERT & Qwen3-0.6B & closed \\
\textbf{scoring} & marker & independent & unknown \\
\textbf{parameters} & 322M & 0.6B & n/a \\
\textbf{items} & 8{,}876 & 888 & 8{,}761 \\
\textbf{latency} & 27.2\,ms & 86.1\,ms & API \\
\textbf{weights} & open & open & closed \\
\bottomrule
\end{tabularx}
\caption{\textbf{The three model arms.} Item counts are each arm's main run: the
full ladder for Laya-ml, a fixed 10\% random subset for Lod-lille, and the English
pool for Jev-1.13. Every arm also ran the $3{,}265$-item capture matrix in full.
Parameter counts are the model cards'. Appendix~\ref{app:subset} bounds the
subset's drift at $0.016$.}
\label{tab:models}
\end{table}

\paragraph{Request.} For the commercial arm the body posted to
\texttt{/api/alpha/decisions} is the following; the instruction string is the
one in Table~\ref{tab:instr}.

\begin{quote}\scriptsize
\begin{alltt}
\{
  "model": "typesafe/jev-1.13",
  "state": \textit{<idiom or carrier sentence>},
  "questions": \{
    "q": \{
      "type": "choice",
      "instructions": \textit{<instruction string>},
      "criteria": \{
        "A": \textit{<gloss>},
        "B": \textit{<gloss>},
        "C": \textit{<gloss>},
        "D": \textit{<gloss>}
      \}
    \}
  \}
\}
\end{alltt}
\end{quote}

The two open arms are called locally with the same four fields: the idiom as
state, the instruction string, and the four option texts keyed \texttt{A}--%
\texttt{D}. The option order is shuffled per item by a generator seeded with the
item id and the experiment seed, so within an experiment the same item receives the
same permutation in both open arms and in every family. The commercial arm's blocks
use their own seeds (Appendix~\ref{app:repro}), so its permutations differ.

\paragraph{Response.}

\begin{quote}\scriptsize
\begin{alltt}
\{
  "type": "choice",
  "choice": "B",
  "confidence": 0.75,
  "probabilities": \{
    "A": 0.00,
    "B": 0.84,
    "C": 0.16,
    "D": 0.00
  \}
\}
\end{alltt}
\end{quote}

We assert on every call that the returned keys are a subset of
\texttt{\{A,B,C,D\}} and that the probabilities sum to $1$ within $10^{-2}$; no
call in any run violated either.

\paragraph{Two confidence signals, and which one we use.} Note that
\texttt{confidence} is not the maximum probability: in the example above it is
$0.75$ while $\max_o p(o) = 0.84$. It is a separate head whose documented job is
to estimate whether the correct answer is present in the option list at all.
Lod-lille ships the same arrangement. We log it as \texttt{conf\_head} and
report it where it matters (\S\ref{sec:confidence}), but every cross-model
number in this paper uses the maximum probability, because it is the only
confidence signal all three arms define identically. Where the two disagree the
head is not the more diagnostic of the pair: on Lod-lille's captured items mean
head confidence is $0.631$ against $0.295$ on correct ones, the same inversion
the top probability shows.

\paragraph{Answer types we did not use.} The interface also accepts
\texttt{score} (a bounded numeric judgment) and \texttt{boolean} questions. We
use \texttt{choice} exclusively and deliberately. A \texttt{score} question
would ask the model to rate a gloss in isolation, which destroys the property
the instrument depends on: that the gold and the literal reading compete inside
a single forward pass, exactly as the psycholinguistic account describes
\citep{cacciari1988comprehension,titone1999compositional}. A \texttt{boolean}
question would do the same and additionally reintroduce a yes/no label whose
name carries meaning of its own, which is the failure
\citet{sun2026typesafe} document. Keeping every experiment in one answer type
also means the five distractor families differ in nothing but their option text.

\begin{table}[t]
\centering\small
\renewcommand{\arraystretch}{1.12}
\begin{tabularx}{\columnwidth}{@{}>{\raggedright\arraybackslash}p{1.35cm}>{\raggedright\arraybackslash}X@{}}
\toprule
\textbf{Arm} & \textbf{Instruction string} \\
\midrule
Bengali &
\bangla{নিচের বাংলা বাগধারাটির প্রকৃত ভাবার্থ কোনটি?} \\
English &
``What is the actual figurative meaning of this idiom?'' \\
\bottomrule
\end{tabularx}
\caption{The two instruction strings. Each is one sentence and is byte-identical
across families, experiments and arms; the Bengali string is the direct equivalent
of the English one. Which string a call receives follows the language of the cell,
on every arm: Bengali cells receive the Bengali string and English cells the
English one, including the open arms' English cells in E4 and in the capture
matrix. The open arms' task configuration records only the Bengali string because
their English cells set the English one in the run code. The matched condition of
Appendix~\ref{app:threeway} therefore shares its instruction across the three
arms, and differs only in the sampled distractors, which the commercial arm drew
under its own seed namespace.}
\label{tab:instr}
\end{table}

\section{Positioning against concurrent work}
\label{app:positioning}

The decision-model literature is a few weeks old at the time of writing, and
several papers describe failures adjacent to ours. This appendix states what
each one establishes, what it does not, and why our design is complementary
rather than duplicative.

\paragraph{Mechanism versus behavior.} \citet{oh2026tug} apply causal tracing
to autoregressive transformers and recover two pathways carrying an idiom's
figurative and literal readings at once: early layers and specific attention
heads promote the figurative interpretation while a direct route keeps the
literal one available, and disambiguating context is integrated from the first
layer onward. That is mechanistic evidence that the competition our \Dlit{}
condition exploits is real inside a model, and it is what makes our outcome
measure interpretable rather than arbitrary: when a decision model places its
mass on the literal gloss, there is a documented sense in which both readings
were present and one won.

The two designs answer different questions, and the difference is the reason
both are needed. Causal tracing requires weight access, token-level activations
and a generative forward pass to intervene on. Two of our three arms release
weights and the third does not, but none of the three emits a token stream to
trace, and the commercial one is reachable only through an HTTP interface. Their method cannot be pointed at
the systems that are currently being deployed as routers and guardrails; ours
can, at the cost of saying nothing about where inside the model the competition
is resolved. They establish \emph{how} the literal route survives; we measure
\emph{how often} it takes over, in which language, and with what confidence
attached.

\paragraph{Option names and option definitions.} \citet{sun2026typesafe} hold a
typed model's option definitions fixed and permute which name is bound to which,
and find decision-flip rates up to 70.4 points higher for semantically loaded
labels than for neutral ones: the model follows the name rather than the
definition it was given. They note that type-safety conceals this, because every
output remains structurally valid.

Our \textsc{o}-series (\S\ref{sec:bagofwords}) manipulates the complementary
half. We hold the option names fixed at \texttt{A}--\texttt{D} and destroy the
internal structure of the text bound to them, chiefly by shuffling words within
each option (truncation is the weaker probe, Appendix~\ref{app:options}), and find
accuracy almost unchanged. Read together, the two results point the same way from
either side: within an option, these models are sensitive to lexical identity, and
on random distractors their decisions do not depend on composed meaning. That falls
short of showing they cannot compose, since composition is not needed on that
condition (Appendix~\ref{app:options}). Neither result alone would establish that, because theirs is consistent
with a model that reads definitions carefully but weights names more heavily,
and ours is consistent with a model that reads definitions as unordered bags.

\paragraph{Context that flips decisions.} \citet{xu2026jevout} append short,
fluent, task-preserving context and redirect 61.4\% of initially correct
decisions, with targeted flip rates of 64.9--73.2\% across systems. Our E3
carrier-sentence condition (\S\ref{sec:bagofwords}) is the benign case of the same
mechanism: embedding an idiom in its gold example sentence \emph{raises}
accuracy by $0.080$ and $0.117$ on the two open arms. The two findings are the
same sensitivity with opposite signs, and the uncomfortable implication is that
a caller cannot tell them apart from the interface. A model whose answer
improves when helpful context is appended is, necessarily, a model whose answer
moves when context is appended at all.

\paragraph{Rejection, and the answer type.} \citet{zhong2026rejection} report the
failure closest to our \textsc{o4} control: asked to choose among candidates that
contain no correct answer, Jev rejects on about $7\%$ of cases, while selecting a
correct candidate almost perfectly when one is present. Two scope conditions
matter. Their items are paired arithmetic problems with an explicit rejection
option, which ours are not, so the agreement is between a reading and a result
rather than a replication. And they find the same model rejects correctly on about
$99\%$ of those cases when asked instead for a Boolean verification of each
candidate in turn. That is a different answer type from the choice type we hold fixed
throughout (Appendix~\ref{app:payload}), and it is a genuine possibility that a
verification-typed call would recover what the choice type does not. Our claim is
about what the choice interface reports, which is the interface these systems are
deployed behind; it is not a claim that the underlying model cannot tell.

\paragraph{Calibration and cascades.} \citet{ibrahim2026evaluating} evaluate
decision models against 19 language models on 18 computational social science
tasks, and report a median deficit of 11.6 F1 at 44 times lower cost, with
confidence calibration \emph{exceeding} most baselines; they recommend
decision models as a cheap first stage that routes uncertain items onward.
\citet{li2026jevjudge} build that system, accepting high-confidence verdicts and
escalating the rest, and exceed a frontier model's accuracy at 41\% of its cost.

We reproduce the favorable calibration for the commercial arm on its own
language (\S\ref{sec:confidence}), so this is not a contradiction. The boundary
we add has two parts. First, in-distribution calibration does not survive the
shifts we apply: on the open arms it fails under a targeted distractor shift, and
on the commercial arm, which stays calibrated under that shift in English, it fails
under the change of language. On the open arms it fails in the direction that costs
a cascade most: on the literal-bearing families the high-confidence tail is where the open
arms are \emph{least} accurate, so a confidence gate forwards exactly the items
that should have been escalated. Second, and independent of any distribution
assumption, the semantics-free control (\S\ref{sec:nosignal}) shows what each
arm reports when the option list provably carries no information. A gate
calibrated on real options has no defense against that case, and a degenerate
option list is an ordinary consequence of an upstream bug or an empty retrieval.

\paragraph{What none of this settles.} All of the above, ours included, is
zero-shot evaluation of base checkpoints. \citet{tang2026audit} review 28 papers
published within nine days of the first widely used release and conclude that
the typed readout has shown no independent accuracy advantage, with the
demonstrated gains in latency, cost and cascade placement. Whether any of these
failure modes survives task-specific fine-tuning is, as far as we can determine,
unmeasured by anyone so far.


\begin{table*}[t]
\centering
\small

\begin{tabularx}{\textwidth}{
    @{}
    c
    >{\raggedright\arraybackslash}X
    >{\centering\arraybackslash}p{1.45cm}
    >{\centering\arraybackslash}p{1.55cm}
    >{\centering\arraybackslash}p{1.55cm}
}
\toprule

&
\textbf{option text, as supplied}
&
\textbf{role}
&
\textbf{Laya-ml $p$}
&
\textbf{Jev-1.13 $p$}
\\
\midrule

A &
\bangla{একগর্ভা; এক সন্তানের জননী; যে নারী একবার মাত্র গর্ভ ধারণ করে} &
\textit{distractor} &
0.00 &
0.02
\\[3pt]

B &
\bangla{ওজন} (\emph{physical weight}) &
\textbf{literal} &
\pL{0.99} &
\pJ{0.15}
\\[3pt]

C &
\bangla{ক্ষমতা, গুরুত্ব, মর্যাদা} &
\textbf{gold} &
0.00 &
\pJ{0.82}
\\[3pt]

D &
\bangla{হালকাভাবে পড়া} &
\textit{distractor} &
0.01 &
0.01
\\

\bottomrule
\end{tabularx}

\caption{\textbf{One item exactly as received by the models.}
The state is the Bengali idiom \bangla{ওজন} (\emph{ojon}), literally
``physical weight,'' with the English equivalent \emph{carry weight}.
The instruction is identical across items and distractor families.
The four option texts are the actual strings reconstructed with the
builder and seed used for the run and checked against the recorded gold
and literal keys. The probabilities are the returned values. Laya-ml
places $0.99$ on the literal option, which denotes the physical weight
of an object, and $0.00$ on the gold figurative gloss; Jev-1.13 places
$0.82$ on the gold option.}
\label{tab:bnitem}

\end{table*}

\section{What each experiment changes}
\label{app:experiments}

Table~\ref{tab:expgrid} states, for each experiment, the one thing that varies
and the things held fixed. Reading down the ``varies'' column follows the
argument: each row isolates one candidate explanation for the failure measured
in E1.

\begin{table}[h]
\centering\small
\setlength{\tabcolsep}{3pt}
\begin{tabular}{llll}
\toprule
& \textbf{Varies} & \textbf{Fixed} & \textbf{Seed} \\
\midrule
\neu{E1} & \neu{distractors} & \neu{idiom, instr., $k$} & \neu{1000} \\
E2 & idiom surface & meaning, options & 2000 \\
E3 & idiom form & options, instr. & 3000 \\
E4 & task language & corpus row, family & 4000 \\
\neu{E5} & \neu{nothing (derived)} & \neu{n/a} & \neu{n/a} \\
E6 & option text & idiom, gold & 6000 \\
E7 & option order & option set & 7000 \\
\bottomrule
\end{tabular}
\caption{The seven interventions. E5 is the calibration and selective-prediction
analysis of \S\ref{sec:confidence}: it issues no calls of its own and is computed
from the E1 outputs. E4's two cells are paired corpus rows rather than the same
strings: the English cell presents the annotator's English equivalent idiom and
the English glosses, so what is held fixed is the row and the family, not the
state (Appendix~\ref{app:construction}). Seeds are the Bengali arms'; the
commercial arm uses a disjoint set (Appendix~\ref{app:repro}) because its blocks
were run in a different order.}
\label{tab:expgrid}
\end{table}

\paragraph{E1, the ladder.} \Drand{} samples three other idioms' figurative
glosses uniformly. \Dsurf{} samples glosses sharing at least one surface token
with the gold. \Dsem{} samples glosses from idioms sharing at least one semantic
tag. \Dlit{} replaces one sampled distractor with the idiom's own expert literal
gloss. \Dhard{} applies all three pressures at once. Duplicates are removed and
the slate is topped up from the random pool if fewer than $k$ distinct options
survive.

\paragraph{E3, perturbing the unit.} E3 runs on the literal-bearing option set,
so its accuracies are comparable to \Dlit{} rather than to \Drand: on the same
1{,}778 items Laya-ml scores $0.176$ under \Dlit{} and $0.353$ under \Drand,
against $0.157$ for the intact idiom here. The $0.176$ and $0.157$ are the same
model on the same items under the same family, and they differ because E1 and E3
draw their random distractors from separate seed namespaces (1000 and 3000), so
the two option lists share the literal gloss and the gold but not the two sampled
distractors. Comparisons are therefore made within an experiment, never across
one. \textsc{v0} is the intact idiom.
\textsc{v1} shuffles its words, \textsc{v2} reverses them, \textsc{v3} deletes
one word at random, \textsc{v4} substitutes one word with a word sampled from
another idiom, and \textsc{v5} replaces the idiom with its gold carrier
sentence. The option list is held byte-identical across all six, so the only
difference reaching the model is the state string.

\paragraph{E6, ablating the options.} \textsc{o0} is the full gloss. \textsc{o1}
truncates every option to its first six words. \textsc{o2} shuffles the words
within each option. \textsc{o3} replaces each gloss with its counterpart in the
other language. \textsc{o4} replaces every option with a bare label carrying no
semantics, which pins accuracy at chance by construction and is the control of
\S\ref{sec:nosignal}.

\section{Why interventions on the option set}
\label{app:why}

This appendix explains the design choice the rest of the paper rests on, for
readers who want the reasoning rather than the result.

A decision model takes a question and a list of candidate options and returns a
probability over exactly those options. It does not sample and it emits no text. Three of the usual explainability routes are
therefore closed or compromised. There is no rationale to inspect and no chain of thought whose
faithfulness could be measured \citep{jacovi2020faithfulness,turpin2023unfaithful}.
Weights are available for two of our arms but not for the commercial one, so no
method that requires opening every model, including probing and causal tracing,
can be applied uniformly across the three \citep{oh2026tug}. And input attribution, the usual
fallback \citep{ribeiro2016lime,lundberg2017shap}, is not only unreliable here
\citep{jain2019attention,adebayo2018sanity,hooker2019roar} but hard to interpret
for this material.

The last point needs stating carefully, because it is what motivates everything
else, and the standard shorthand for it is too strong. Idioms are not uniformly
non-compositional. \citet{nunberg1994idioms} argue the opposite of the slogan
usually attributed to them: idioms vary in how far their parts can be mapped onto
components of the figurative reading, and many are \emph{idiomatically
decomposable}, with \emph{spill the beans} distributing its meaning across
\emph{spill} and \emph{beans} in a way that \emph{kick the bucket} does not.
Psycholinguistic work finds decomposability measurable and consequential for
processing \citep{gibbs1989kick,titone1999compositional}.

What follows is weaker than ill-posedness but enough for our purposes. A Shapley
value over the tokens of an idiom is always computable, and for a
non-decomposable expression there is no reason to expect the resulting per-token
shares to correspond to anything linguistically real: the quantity is defined by
the attribution game rather than by a decomposition the expression actually
supports, and decomposability is a property of the individual idiom that we do not
have annotated. The numbers will still arrive, and they will still read as an
explanation. We therefore do not build on them, and intervene on the interface
instead, which requires no assumption about where an idiom's meaning sits.

What remains is the interface itself. The caller controls the question and the
option list and observes a probability vector. So we intervene there. The
corpus supplies, for every idiom, both a gold figurative gloss and an expert
literal gloss, which means a counterfactual distractor is already annotated and
we create no new labels: dropping the literal gloss into the option list turns a
benchmark item into a targeted probe of whether the expression was read as an
idiom or as its words. Holding the item, the instruction and $k$ fixed while
varying only the distractor family makes the comparison within-item and
within-model, so a difference between two families cannot be explained by item
difficulty, by the prompt, or by anything about the model that does not change
between the two calls.

The same logic produces the other five interventions; E5 is an analysis of E1's
outputs rather than an intervention. If the failure were about
the idiom's surface form rather than its meaning, E2 would show it
(Appendix~\ref{app:surface}). If the model
held the expression as a stored unit, breaking the unit in E3 would hurt. If the
problem were figurativeness rather than the language, E4 would not move. If the
model read options as sentences rather than as bags of words, E6 would collapse
under scrambling. If the scorer were order-independent as documented, E7 would
return the same gloss under all 24 permutations. Each experiment is an attempt
to explain the E1 result away, and the paper reports which attempts succeed.

\section{The full distractor ladder}
\label{app:ladder}

Table~\ref{tab:fullladder} gives every family for every arm. Two caveats on the
commercial column: its numbers are English, and its \Dsurf{} and \Dsem{} rows
run on the larger 8{,}761-item English pool rather than the 3{,}265-item
literal pool, because those families need no literal gloss.

\paragraph{What the ladder is for.} The five families form an ordered difficulty
scale in which only the distractor construction changes. \Drand{} draws three
other idioms' figurative glosses uniformly, so a wrong answer requires confusing
two unrelated meanings. \Dsurf{} draws glosses belonging to idioms that share a
word with the target, which makes the distractors lexically adjacent to the state
but not to the gold. \Dsem{} draws from idioms sharing a topical tag, which makes
them semantically adjacent to the gold. \Dlit{} replaces one distractor with the
target's own expert literal gloss. \Dhard{} combines the literal gloss with one
surface and one semantic neighbor. Reading down the first four rows therefore
separates three different kinds of hardness, and the point of the ladder is that
they are not interchangeable.

\paragraph{The literal family is not merely the hardest.} Measured as a drop from
\Drand{}, the lexical and semantic neighbors cost Laya-ml $0.029$ and $0.034$,
and Lod-lille $0.072$ and $0.038$. The literal gloss costs them $0.216$ and
$0.418$: roughly six to eleven times as much as either neighbor family, from a
single substituted option. A benchmark that stopped at \Dsem{}, which is where
most distractor-hardness work stops, would have recorded these two models as
moderately sensitive to distractor difficulty and would have missed the effect
entirely. That is the argument for the family being the independent variable
rather than a robustness check.

\paragraph{The commercial column does something different.} Jev-1.13's ordering is
not the same shape. Its worst family in English is \Dsem{} at $0.829$, and the
literal family at $0.931$ is \emph{easier} for it than the semantic one, which
inverts the open arms' ordering. Those two rows come from different pools, so that
comparison is indicative; the paired version is on the $3{,}265$ items both
\Drand{} and \Dlit{} share, where the literal gloss costs it only $0.037$
($0.968$ to $0.931$) against $0.216$ and $0.418$ on the open arms. Read alone, that looks like a model which has
simply solved the literal distractor. Read against the Bengali column of
Table~\ref{tab:matrix}, where the same model on the same construction is captured
at $0.304$, it reads instead as a model for which the literal distractor is easy
\emph{in English}. Only the matrix distinguishes the two, which is why we report
the English ladder and the matrix together rather than either alone.

\paragraph{Why \Dhard{} adds little.} \Dhard{} keeps \Dlit{}'s literal gloss and
replaces its two random distractors with harder neighbors, yet costs only a further
$0.007$ and $0.003$ on the open arms. Once the literal gloss is present it absorbs essentially all of the
available error: the remaining two slots could be filled with anything without
changing the outcome much. This is also why we report \LAR{} on both families and
treat \Dlit{} as the headline, since it is the cleaner single-variable
manipulation.

\begin{table}[t]
\centering\small
\renewcommand{\arraystretch}{1.12}
\begin{tabularx}{\columnwidth}{@{}>{\raggedright\arraybackslash}X>{\centering\arraybackslash}p{1.35cm}>{\centering\arraybackslash}p{1.35cm}>{\centering\arraybackslash}p{1.35cm}@{}}
\toprule
\textbf{Family} & \textbf{Laya-ml} & \textbf{Lod-lille} & \textbf{Jev-1.13} \\
& bn & bn & en \\
\midrule
\Drand & 0.367 & 0.484 & 0.968 \\
\Dsurf & 0.338 & 0.412 & 0.915 \\
\Dsem  & 0.333 & 0.446 & 0.829 \\
\Dlit  & 0.151 & 0.066 & 0.931 \\
\Dhard & 0.144 & 0.063 & 0.854 \\
\midrule
\LAR{} \Dlit  & 0.644 & 0.885 & 0.047 \\
\LAR{} \Dhard & 0.630 & 0.855 & 0.043 \\
\bottomrule
\end{tabularx}
\caption{Accuracy by distractor family, and literal attraction on the two
families that contain a literal gloss. Chance is $0.250$. The commercial column
is English and is the reason \S\ref{sec:capture} runs the matrix: compare it
with the Bengali column of Table~\ref{tab:matrix}, where the same model is
captured at $0.304$.}
\label{tab:fullladder}
\end{table}

\begin{figure*}[t]
\centering
\includegraphics[width=0.72\linewidth]{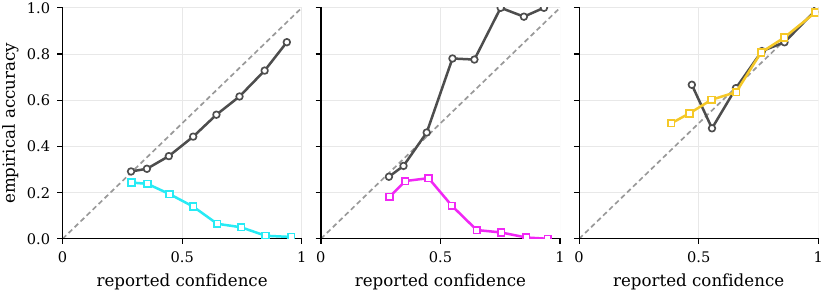}
\caption{Reliability, 15 equal-width bins. Panels left to right:
\sw{cLaya}Laya-ml, \sw{cLod}Lod-lille, \sw{cJev}Jev-1.13. \swK circles are the
random-distractor condition, squares in each panel's own color the $+$literal
one, and the gray dashed diagonal perfect calibration. Under \Dlit{} the two open
models' curves bend \emph{downward}: the model is most wrong where it reports
being most sure. Jev-1.13, in English, stays on the diagonal.}
\label{fig:reliability}
\end{figure*}

\section{Selective prediction in detail}
\label{app:selective}

Selective prediction asks what happens if the caller answers only the items the
model is most confident about. A useful confidence signal makes accuracy rise as
coverage falls. Table~\ref{tab:riskcov} reports accuracy at low coverage for the
two open arms, on the random-distractor condition and on the literal-bearing
one.

\begin{table}[t]
\centering\small
\renewcommand{\arraystretch}{1.12}
\begin{tabularx}{\columnwidth}{@{}>{\raggedright\arraybackslash}p{1.3cm}*{4}{>{\centering\arraybackslash}X}}
\toprule
& \multicolumn{2}{c}{\textbf{Laya-ml}} & \multicolumn{2}{c}{\textbf{Lod-lille}} \\
\cmidrule(lr){2-3}\cmidrule(lr){4-5}
\textbf{Coverage} &
\laya{\textbf{\Drand}} &
\laya{\textbf{\Dlit}} &
\lod{\textbf{\Drand}} &
\lod{\textbf{\Dlit}} \\
\midrule
0.05 & \laya{0.669} & \laya{0.007} & \lod{0.977} & \lodh{0.000} \\
0.10 & \laya{0.597} & \laya{0.007} & \lod{0.989} & \lodh{0.000} \\
0.20 & \laya{0.528} & \laya{0.023} & \lod{0.899} & \lodh{0.000} \\
0.35 & \laya{0.470} & \laya{0.047} & \lod{0.785} & \lodh{0.000} \\
\midrule
full & \laya{0.367} & \laya{0.151} & \lod{0.484} & \lod{0.066} \\
\bottomrule
\end{tabularx}
\caption{\textbf{Accuracy under a confidence gate.}
The two model arms are named in the column-group headers and also carry their
usual tints, \sw{cLaya}cyan for Laya-ml and \sw{cLod}magenta for Lod-lille, so
the table reads without color. On random distractors, both arms behave as a caller would
hope: restricting to the most confident 5\% roughly doubles accuracy.
On the literal-bearing condition, the curve inverts. Laya-ml falls from
$0.151$ at full coverage to $0.007$ in its most confident 5\%, while
Lod-lille answers \textbf{every one} of its most confident 35\% incorrectly.
Gating on confidence is therefore worse than not gating, and worse than
gating at random.}
\label{tab:riskcov}
\end{table}

This is the result with the most direct operational consequence. A cascade that
forwards high-confidence answers and escalates the rest
\citep{li2026jevjudge,ibrahim2026evaluating} forwards precisely the items on
which these models are wrong, and escalates the ones they would have got right.

\paragraph{How to read the table.} Each row fixes a coverage level and reports
accuracy on the retained items, so a row is not a subset of the row above it in
any model-independent way: the items retained at $0.05$ coverage are whichever
items \emph{that} model ranked highest, which differ between the two arms and
between the two families. The comparison the table supports is therefore vertical
within a column, not horizontal across them. The quantity to watch is the sign of
the gradient. In the two \Drand{} columns it is the sign a caller expects, with
one small reversal at the top of Lod-lille's ($0.977$ at $5\%$, $0.989$ at $10\%$),
and the magnitude is substantial: Lod-lille reaches $0.977$ in its most confident
twentieth against $0.484$ at full coverage. In the two \Dlit{} columns it is
reversed, and the reversal is monotone rather than noisy, which is what makes it
an inversion of the mechanism rather than a weak signal. The four zeros in the
last column are literal: across the $311$ items making up Lod-lille's most
confident $35\%$ under \Dlit{}, the model is correct on none.

\paragraph{Calibration without binning.} Table~\ref{tab:calib} reports the same
distributions under three metrics. The point of giving three is that each fails
differently. ECE with equal-width bins is the field standard and is also the
metric \citet{nixon2019measuring} show to be sensitive to the binning: with 15
equal-width bins most of the mass can land in one or two bins, and an estimator
averaging over near-empty bins can move with the bin count. Replacing them with
15 equal-mass bins, which is their adaptive variant, changes nothing material
here: the largest shift across the nine cells is $0.011$, and the ordering of the
arms is identical. So the calibration failure is a property of the
distributions, not of our bin edges.

The Brier score is the stronger statement because it needs no bins at all. It is
strictly proper, so no reporting strategy beats reporting one's true beliefs, and
it has a fixed reference point: a model that returns $0.25$ on all four options
scores exactly $0.750$. Under \Drand{} both open arms sit below that reference,
which is what a model carrying some signal should do. Under \Dlit{} both rise
above it, to $1.030$ and $1.391$. This is worth stating plainly because it is
stronger than miscalibration: on the literal-bearing condition, a caller
integrating these models' full reported distributions is worse off than a caller
who replaced them with a flat prior. Jev-1.13 in English stays far below the
reference throughout, and the one cell where it approaches a visible calibration
cost is \Dhard{} ($0.220$, ECE $0.047$), which with \Dlit{} is one of the two
conditions containing the literal gloss.

\begin{table}[t]
\centering\small
\renewcommand{\arraystretch}{1.12}
\setlength{\tabcolsep}{3.2pt}
\begin{tabular}{@{}l*{4}{c}@{\hspace{5pt}}cc}
\toprule
\textbf{Family} & \textbf{acc} & \textbf{ECE} & \textbf{ACE} & \textbf{Brier} &
\textbf{AUGRC} & \textit{rand} \\
\midrule
\multicolumn{7}{@{}l}{\laya{\emph{Laya-ml} ($n=8{,}876$)}} \\
\Drand & 0.367 & 0.071 & 0.073 & 0.728 & 0.294 & \textit{.316} \\
\Dlit  & 0.151 & 0.385 & 0.385 & \textbf{1.030} & \textbf{0.448} & \textit{.425} \\
\Dhard & 0.144 & 0.389 & 0.389 & \textbf{1.030} & \textbf{0.450} & \textit{.428} \\
\addlinespace
\multicolumn{7}{@{}l}{\lod{\emph{Lod-lille} ($n=888$)}} \\
\Drand & 0.484 & 0.073 & 0.084 & 0.608 & 0.193 & \textit{.258} \\
\Dlit  & 0.066 & 0.679 & 0.681 & \textbf{1.391} & \textbf{0.491} & \textit{.467} \\
\Dhard & 0.063 & 0.673 & 0.674 & \textbf{1.390} & \textbf{0.490} & \textit{.468} \\
\addlinespace
\multicolumn{7}{@{}l}{\jev{\emph{Jev-1.13}, English ($n=3{,}265$)}} \\
\Drand & 0.968 & 0.005 & 0.003 & 0.048 & 0.002 & \textit{.016} \\
\Dlit  & 0.931 & 0.014 & 0.013 & 0.107 & 0.010 & \textit{.035} \\
\Dhard & 0.854 & 0.047 & 0.047 & 0.220 & 0.034 & \textit{.073} \\
\bottomrule
\end{tabular}
\caption{\textbf{Three calibration metrics on the same recorded
distributions.} \swL Laya-ml, \swD Lod-lille, \swJ Jev-1.13. ECE uses 15
equal-width bins, ACE the adaptive equal-mass variant of
\citet{nixon2019measuring}; the largest gap between them across these nine cells
is $0.011$, so the ECE figures elsewhere in the paper are not artifacts of the
binning. Brier is the multiclass score and needs no bins; bold marks cells above $0.750$,
its value for a uniform four-way distribution, where the model's reported
distribution is less useful to a caller than no preference at all. AUGRC is the
generalized risk-coverage area of \citet{traub2024selective}, which does not
condition risk on the accepted set and so does not over-weight high-confidence
failures; the \textit{rand} column is its analytic random baseline, half the
error rate, and bold marks the AUGRC cells that are worse than it. No new model calls were
made; every metric here is a re-reading of the released per-item probabilities.}
\label{tab:calib}
\end{table}

\section{Stratified results}
\label{app:strata}

Literal capture is not concentrated in rare or obscure idioms, which would be
the comfortable explanation. Splitting Laya-ml's \Dlit{} condition by the
corpus's own frequency annotation gives \LAR{} $0.671$ on \emph{very common}
idioms ($n = 1{,}146$), $0.646$ on \emph{common} ($n = 6{,}200$), $0.615$ on
\emph{rare} ($n = 1{,}507$) and $0.565$ on \emph{very rare} ($n = 23$). If
anything the effect is strongest on the most familiar expressions, which is the
opposite of a long-tail story and consistent with these being the idioms whose
literal readings are most available.

The gradient is monotone across the three populated strata and spans $0.056$,
which is small against the effect itself but points the wrong way for a
familiarity account. We would not build on the \emph{very rare} cell: at $n=23$
its interval is wide enough to be consistent with anything, and we report it only
so the stratification is complete.

The corpus's \texttt{cultural\_significance} field is unusable as a stratifier:
10{,}337 of 10{,}359 rows carry the same value. We mention this because it is the
field a reader would most want stratified, and its absence is a property of the
resource rather than a choice of ours.


\begin{table*}[t]
\centering
\small
\renewcommand{\arraystretch}{1.15}

\begin{tabularx}{\textwidth}{
    @{}
    >{\raggedright\arraybackslash}p{3.45cm}
    >{\centering\arraybackslash}p{1.35cm}
    >{\raggedright\arraybackslash}X
}
\toprule

\textbf{manipulation}
&
\textbf{$p(\text{gold})$}
&
\textbf{string sent to the model}
\\
\midrule

\multicolumn{3}{@{}l}{
    \textbf{E3: the state changes}
    \quad
    \textit{idiom: \bangla{অগাধ জলের মাছ}
    (\emph{Still waters run deep})}
}
\\[1pt]

\textit{intact}
&
\textbf{0.21}
&
\hlL{\bangla{অগাধ জলের মাছ}}
\\

\textit{word shuffle}
&
\textbf{0.19}
&
\bangla{মাছ জলের অগাধ}
\\

\textit{reversed}
&
\textbf{0.19}
&
\bangla{মাছ জলের অগাধ}
\\

\textit{delete one word}
&
\textbf{0.24}
&
\bangla{অগাধ জলের}
\\

\textit{substitute one word}
&
\textbf{0.24}
&
\bangla{অগাধ জলের ফ্যা}
\\

\textit{in its carrier sentence}
&
\textbf{0.07}
&
\bangla{সে অগাধ জলের মাছ, কখনো বুঝতে পারি না তার পরিকল্পনা কী।}
\\[4pt]

\midrule

\multicolumn{3}{@{}l}{
    \textbf{E6: the options change}
    \quad
    \textit{gold option rewritten; the other three are rewritten identically}
}
\\[1pt]

\textit{full gloss}
&
&
\hlL{\bangla{খুব চালাক লোক; সুচতুর ব্যক্তি; যার মনোভাব বোঝা যায় না}}
\\

\textit{truncated to 6 words}
&
&
\bangla{খুব চালাক লোক সুচতুর ব্যক্তি যার}
\\

\textit{words scrambled}
&
&
\bangla{খুব লোক সুচতুর চালাক যায় ব্যক্তি না যার বোঝা মনোভাব}
\\

\textit{English gloss}
&
&
A very clever person; a shrewd individual; someone whose mindset or
intentions are hard to fathom
\\

\textit{semantics removed}
&
&
\hlG{\bangla{অর্থ} 1}
\\

\bottomrule
\end{tabularx}

\caption{\textbf{What the interventions do to real Bengali text.}
The table shows the exact strings used for the idiom
\bangla{অগাধ জলের মাছ} (\emph{still waters run deep}, literally
``a fish of deep water''); the tinted string in each block is the unmodified one
the rest of that block is derived from. \emph{E3} changes the state supplied to
the model; the reported probabilities are Laya-ml's, on the gold figurative
gloss. E6 rewrites the options rather than the state, so a single
$p(\text{gold})$ is not defined for its rows and the column is empty there. Shuffling and reversing the
expression leave the probability essentially unchanged, whereas the
carrier sentence changes it. \emph{E6} changes the option text while
holding its role fixed; after the full gloss, the rows truncate it, scramble its
words, translate it, and finally remove its content, ending with the semantics-free
control.
The other three options undergo the same transformation as the gold
option. The E6 tokenizer splits on punctuation and rejoins on spaces, so the
semicolons of the full gloss do not survive into the truncated and scrambled
rows; the scrambled row is a strict permutation of the full gloss's ten tokens
and can be checked as one.}

\label{tab:bnmanip}
\end{table*}

\section{Output distributions}
\label{app:dists}

The three arms do not merely differ in accuracy; they place their probability
mass very differently, and the shape of that mass is what makes the confidence
results of \S\ref{sec:confidence} possible. Table~\ref{tab:dist} summarizes the
returned vectors.

\begin{table}[t]
\centering\small
\renewcommand{\arraystretch}{1.12}
\begin{tabularx}{\columnwidth}{@{}ll*{4}{>{\centering\arraybackslash}X}@{}}
\toprule
& & \textbf{Mean} & \textbf{Med.} & $\mathbf{>}\textbf{.95}$ & \textbf{Ent.} \\
\midrule
\multirow{2}{*}{Laya} &
\Drand & 0.438 & 0.408 & 0.001 & 0.892 \\
& \Dlit & 0.536 & 0.479 & 0.039 & 0.778 \\
\multirow{2}{*}{Lod} &
\Drand & 0.447 & 0.391 & 0.002 & 0.881 \\
& \Dlit & 0.745 & 0.819 & 0.171 & 0.497 \\
\multirow{2}{*}{Jev} &
\Drand & 0.967 & 1.000 & 0.852 & 0.068 \\
& \Dlit & 0.931 & 0.990 & 0.708 & 0.131 \\
\bottomrule
\end{tabularx}
\caption{Distribution of the returned maximum probability, and normalized
entropy of the full vector ($1=$ uniform). The literal-bearing family makes
both open arms \emph{sharper}, not flatter: Lod-lille's mean peak rises from
$0.447$ to $0.745$ and the share of near-deterministic answers from $0.002$ to
$0.171$, while its accuracy falls from $0.484$ to $0.066$. The model is not
confused by the literal gloss; it is convinced by it.}
\label{tab:dist}
\end{table}

That last point is worth stating separately, because it is the mechanism behind
the inverted reliability curves. A model that found the literal option merely
distracting would spread mass and lose accuracy together. Both open arms do the
opposite: adding the literal gloss \emph{concentrates} the distribution onto it.
On 21 items Laya-ml returns $p = 1.000$ on the literal gloss and $p = 0.000$ on
the gold, to three decimal places.

\paragraph{How to read the entropy column.} Normalized entropy is the entropy of
the returned four-way vector divided by $\log 4$, so $1$ is uniform and $0$ is a
point mass. It is reported alongside the maximum because the two can move
independently and the pair identifies the shape. A model that is uncertain between
two options and has ruled out the other two has a high maximum and a middling
entropy; a model hedging across all four has a low maximum and a high entropy.
Both open arms under \Drand{} sit near $0.89$, which is a mild preference over an
almost flat vector: at $0.44$ mean maximum they are closer to guessing than their
accuracy suggests. Under \Dlit{} Lod-lille's entropy falls to $0.497$ while its
accuracy falls to $0.066$, so it moves from hedging to committing at precisely the
point where committing is wrong.

Jev-1.13 is a different instrument. Its entropy is $0.068$ under \Drand{} and
$85.2\%$ of its answers exceed $0.95$, so it is nearly a hard classifier that
happens to expose probabilities. That matters for how its calibration figures
should be read: a model returning $1.000$ almost always has little room to be
miscalibrated in the ECE sense while it is accurate, and correspondingly no room
to signal doubt when it is not. Its low ECE in Table~\ref{tab:conf} and its $0.815$
confidence on the semantics-free control (\S\ref{sec:nosignal}) are the same
property seen from two sides.

\paragraph{Worked examples.} Table~\ref{tab:bnexamples} gives five items in the
original script with the annotators' own glosses and English equivalents. They are
drawn from the extreme tail rather than sampled, items where the mass on the
literal gloss is at or near $1.000$, and we flag the selection because the tail is
what an aggregate \LAR{} of $0.644$ does not convey. All five also come from the
$2{,}480$ items of \S\ref{sec:construction} on which neither lexical cue is
available, so none of them is a case where the idiom and its literal gloss are
the same string or where the literal option is the only one in Latin script. What
they add is the \emph{kind} of expression involved: \bangla{জলের দাম}
(\emph{joler dam}), the price of water, means dirt cheap, and \bangla{দু কথা}
(\emph{du kotha}), two words, means a trifling remark. These are everyday
constructions, and the literal reading the model selects is in every case the
annotators' own rather than one we wrote.


\begin{table*}[t]
\centering
\small
\renewcommand{\arraystretch}{1.15}
\begin{tabularx}{\textwidth}{
    @{}
    >{\raggedright\arraybackslash}p{2.6cm}
    >{\raggedright\arraybackslash}p{3.5cm}
    >{\raggedright\arraybackslash}X
    >{\raggedright\arraybackslash}p{2.4cm}
    >{\centering\arraybackslash}p{1.25cm}
}
\toprule
\textbf{idiom} & \textbf{expert literal gloss} & \textbf{gold figurative gloss} &
\textbf{English equivalent} & \textbf{$p(\text{literal})$} \\
\midrule
\bangla{জলের দাম}
& \bangla{জলের মূল্য} \newline \emph{(the price of water)}
& \bangla{অতি সস্তাদর} \newline \emph{(extremely cheap)}
& \emph{dirt cheap} & \cellcolor{fLaya}\textbf{0.99} \\

\bangla{অক্ষর পরিচয়}
& \bangla{অক্ষর বা অক্ষরমালার জ্ঞান বা পরিচয়} \newline \emph{(knowledge of letters or the alphabet)}
& \bangla{সামান্যতম জ্ঞান} \newline \emph{(the slightest knowledge)}
& \emph{not know one's ABCs} & \cellcolor{fLaya}\textbf{0.99} \\

\bangla{দু কথা}
& \bangla{দুটি কথা} \newline \emph{(two words)}
& \bangla{সামান্য কথা; অল্প ও সাধারণ কথা} \newline \emph{(a trifling remark; a brief and ordinary exchange)}
& \emph{a word or two} & \cellcolor{fLaya}\textbf{0.99} \\

\bangla{রাম-লক্ষণ}
& \bangla{রাম এবং লক্ষ্মণ} \newline \emph{(Rama and Lakshmana, two brothers)}
& \bangla{ভ্রাতৃভক্ত দুই ভাই} \newline \emph{(two devoted brothers)}
& \emph{blood brothers} & \cellcolor{fLaya}\textbf{1.00} \\

\bangla{পেঁয়াজ পয়জার দুইই হওয়া}
& \bangla{পেঁয়াজ এবং জুতা উভয়ই হওয়া} \newline \emph{(to be both the onion and the shoe)}
& \bangla{মার ও গালাগালি খাওয়া} \newline \emph{(to take both a beating and abuse)}
& \emph{catch hell} & \cellcolor{fLaya}\textbf{0.98} \\
\bottomrule
\end{tabularx}
\caption{\textbf{Items where Laya-ml places almost all probability on the literal
gloss.} All five are drawn from the $2{,}480$ items on which none of the
lexical construction cues of \S\ref{sec:construction} is available: each idiom is
multiword, its literal gloss does not contain the idiom string, and the gloss
carries no English parenthetical, so no option in the list carries Latin script.
Each idiom does still share a content word with its literal gloss, the hook that
Table~\ref{tab:hook} tests and finds does not carry capture. Glosses and English equivalents are
the corpus's own; the English in parentheses is ours, for readability. Both the
literal and the figurative gloss were in the option list, and the last column is
the mass placed on the literal one under \Dlit{} on the full Bengali run. On every
one of these the model gave the gold gloss at most $0.011$.}
\label{tab:bnexamples}
\end{table*}

\section{The capture matrix}
\label{app:matrix}

Table~\ref{tab:matrix} gives the six cells of \S\ref{sec:capture} with the
accuracy column alongside \LAR; Table~\ref{tab:matrixci} gives the bootstrap
intervals on those \LAR{} values and the paired within-model language effect.
All cells are $n = 3{,}265$ and were built by one function from seed 21000.
The language effect is positive on every arm and the intervals on the two
language cells are disjoint in every case.
Jev-1.13's English \LAR{} here agrees with its English-ladder value to four
decimal places, $0.0472$, which is why we treat the two constructions as
measuring the same thing. Its accuracy in that cell is $0.926$ against $0.931$ in
Table~\ref{tab:fullladder}: same model, same items, same family, but the two runs
draw their two sampled distractors from different seed namespaces (12000 and
21000), so the option lists share the gold and the literal gloss and differ in the
rest. Capture is insensitive to that and accuracy is mildly sensitive, which is
what one would expect.

\begin{figure}[t]
\centering
\includegraphics[width=\columnwidth]{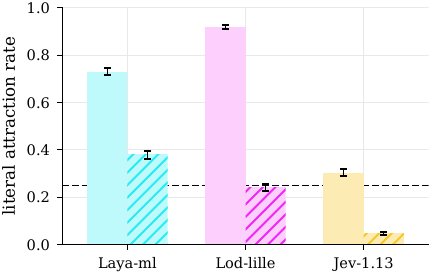}
\caption{Literal attraction by model and task language, all six cells on the same
3{,}265 items (error bars 95\% bootstrap CIs; dashed line the four-option chance
level). \swL Laya-ml, \swD Lod-lille, \swJ Jev-1.13; solid bars Bengali, hatched
bars English. The Bengali ordering does not follow capability, and Jev-1.13's
English value matches the $0.047$ of the English ladder. The cells are in
Table~\ref{tab:matrix} and their intervals in Appendix~\ref{app:matrix}.}
\label{fig:matrix}
\end{figure}

\paragraph{Why this table is the one we defend hardest.} Everywhere else in the
paper a comparison moves one thing and holds the rest fixed within one arm. Here
the design is a full crossing, so the same $3{,}265$ items, the same distractor
draw and the same seed produce all six cells, and both margins are readable. The
language margin is large and uniform in sign: every arm is more captured in
Bengali, by $+0.352$, $+0.677$ and $+0.256$. The capability margin is not even
monotone. Lod-lille, which beats Laya-ml by $+0.133$ on random distractors over
the 888 items they share (Table~\ref{tab:ladder}), is the \emph{most} captured arm
in Bengali, and the strongest arm is the least captured in both languages. A single ordering of
these three models therefore cannot predict their capture, while knowing the test
language can.

\paragraph{Reading accuracy against \LAR{}.} The two columns per language are not
redundant. Accuracy counts items answered correctly; \LAR{} counts items where the
chosen option was specifically the idiom's own literal gloss, which is one of the
three available errors. The gap between them is the residual error on other
distractors. For Lod-lille in Bengali, accuracy $0.044$ and \LAR{} $0.918$ leave
$0.038$ for everything else, so the literal gloss accounts for almost the whole
error budget rather than merely enlarging it. For Jev-1.13 in Bengali, accuracy
$0.536$ and \LAR{} $0.304$ leave $0.160$, so even its errors are less
literal-dominated. This is the sense in which capture is a specific failure and
not a restatement of accuracy.

\paragraph{What the table cannot settle.} Three things. The pool is the $36.8\%$ of
the corpus that carries a usable English gloss, so the levels are pool levels
(Appendix~\ref{app:construction}). The two language cells are paired entries rather
than translations, and are asymmetric in construction in four ways we enumerate in
the same appendix. And the matrix covers the literal condition only; we did not
cross the other four families with language, so we cannot say whether the language
effect is specific to literal capture or general to distractor difficulty. The
first of these affects levels, the second we test in part, since the recorded
outputs reach two of the four asymmetries, and the third is simply unrun.

\begin{table}[t]
\centering\small
\renewcommand{\arraystretch}{1.12}
\begin{tabularx}{\columnwidth}{@{}l>{\centering\arraybackslash}X>{\centering\arraybackslash}X>{\centering\arraybackslash}X>{\centering\arraybackslash}X@{}}
\toprule
& \multicolumn{2}{c}{\textbf{Bengali}} & \multicolumn{2}{c}{\textbf{English}} \\
\cmidrule(lr){2-3}\cmidrule(lr){4-5}
& \textbf{acc} & \textbf{\LAR} & \textbf{acc} & \textbf{\LAR} \\
\midrule
Laya-ml  & 0.127 & 0.731 & 0.417 & \textbf{0.379} \\
Lod-lille & 0.044 & \textbf{0.918} & 0.684 & 0.241 \\
Jev-1.13  & 0.536 & 0.304 & 0.926 & 0.047 \\
\bottomrule
\end{tabularx}
\caption{The literal condition as a model $\times$ language matrix, all six
cells measured on the same 3{,}265 items with one construction and one seed
($n=3{,}265$ per cell; 95\% CIs in Appendix~\ref{app:matrix}). Bold marks the
highest \LAR{} per language. Every arm is far more captured in Bengali than in
English, and the Bengali ordering does not follow capability: Jev-1.13, the
strongest arm, is the least captured in both languages, while Lod-lille, stronger
than Laya-ml on random distractors (Table~\ref{tab:ladder}), is the more captured
of the two in Bengali.}
\label{tab:matrix}
\end{table}

\begin{table}[t]
\centering\small
\renewcommand{\arraystretch}{1.12}
\setlength{\tabcolsep}{4pt}
\begin{tabularx}{\columnwidth}{@{}>{\raggedright\arraybackslash}X
  >{\centering\arraybackslash}p{1.75cm}
  >{\centering\arraybackslash}p{1.75cm}
  >{\centering\arraybackslash}p{1.05cm}}
\toprule
& \multicolumn{2}{c}{\textbf{\LAR{} 95\% CI}} & \\
\cmidrule(lr){2-3}
& \textbf{Bengali} & \textbf{English} & $\boldsymbol{\Delta}$ \\
\midrule
Laya-ml   & $[.716,.746]$ & $[.363,.397]$ & $+0.352$ \\
Lod-lille & $[.909,.928]$ & $[.227,.256]$ & $+0.677$ \\
Jev-1.13  & $[.288,.319]$ & $[.040,.055]$ & $+0.256$ \\
\bottomrule
\end{tabularx}
\caption{Bootstrap intervals on the \LAR{} cells of Table~\ref{tab:matrix} and
the paired within-model language effect. Intervals are 2{,}000-resample
percentile intervals on the same $3{,}265$ items; $\Delta$ is the
Bengali-minus-English difference on those paired items, McNemar
$p < 10^{-100}$ in all three rows.}
\label{tab:matrixci}
\end{table}

\section{How the two language cells were built}
\label{app:construction}

The capture matrix is the result we lean on hardest, so this appendix states
exactly how its two language cells were constructed, lists every asymmetry
between them that we could find, and reports the tests of the two that the
existing outputs can reach.

\paragraph{Where the glosses come from.} Neither literal gloss is a translation
produced by us or by a machine. The corpus annotates one field per idiom,
\texttt{literal\_meaning}, written in Bengali with an English parenthetical
alongside it, so a single expert-authored field carries both readings. The Bengali
cell uses that field as written. The English cell uses the parenthetical alone,
recovered by a regular expression that accepts a parenthesized span only if it
contains at least three ASCII letters and is at least $60\%$ ASCII alphabetic,
which excludes Bengali asides and bare numerals. Of $8{,}876$ items, $3{,}391$
carry such a parenthetical, $3{,}352$ of those also carry an English equivalent
idiom, and $3{,}265$ survive a final cut requiring the English literal gloss to be
lexically distinct from the English gold gloss (Jaccard $< 0.34$). That pool of
$3{,}265$ is the matrix. The provenance matters because a machine-translated
English gloss would make the English cell a measurement of the translator; an
annotator-written one does not.

\paragraph{The four asymmetries.} We can state these precisely, which is better
than asserting the cells are matched.

\emph{(i) The state differs in kind.} The Bengali state is the idiom itself. The
English state is the annotator's nearest English idiom, taken from the
\texttt{similar\_in\_english} field, not a translation of the Bengali one. The
cells are therefore paired entries, and we use that phrase in the paper rather
than ``identical items''.

\emph{(ii) The Bengali literal option is bilingual.} Because the Bengali cell uses
the annotated field as written, and pool membership requires that field to contain
an English parenthetical, every one of the $3{,}265$ Bengali literal options
contains its own English gloss inside it, averaging $44$ characters against $24$
for the English literal option. This is the most direct of the four, because it
supplies a cue that has nothing to do with reading the expression: no Bengali
figurative gloss in the corpus contains any Latin character, so whenever the
literal option carries a parenthetical it is the \emph{only} option in the list
written partly in Latin script. It is also usually the longer of the two glosses,
though not reliably the longest of the four: that holds on $40.2\%$ of items, and
\S\ref{sec:construction} shows length does not predict capture. ``Select the option
containing English'' would reproduce the Bengali capture numbers for all three
arms and would also predict the English cell showing less capture, because there
the cue is absent. We test it below.

\emph{(iii) The distinctness cut applies to one side only.} The Jaccard
$< 0.34$ filter between gold and literal gloss was applied in English and not in
Bengali. Measured after the fact, the filter turns out not to bind: mean gold-versus-literal
Jaccard is $0.029$ in Bengali against $0.052$ in English, so the
Bengali side is the more separated of the two and would have passed the cut almost
everywhere. This asymmetry is procedural rather than consequential.

\emph{(iv) The literal gloss echoes the state more often in Bengali.} With (ii),
this is one of the two that could plausibly manufacture the result. The Bengali literal gloss shares at
least one content word with the Bengali idiom on $56.2\%$ of pool items; the English
literal gloss shares one with the English equivalent idiom on $25.1\%$. A model
doing order-free lexical matching on the option side, which
\S\ref{sec:bagofwords} shows all three partly are, has a surface hook into the
Bengali literal option that it lacks in English.

\paragraph{Testing asymmetry (iv).} The existing per-item outputs are enough to
test it without running anything new. Split the pool on whether the Bengali literal
gloss shares a content word with the idiom and read \LAR{} in each stratum
(Table~\ref{tab:hook}, upper block). If the hook were driving capture, the
overlap stratum would be far more captured in every arm. It is not: the effect is
$+0.026$ for Laya-ml, $+0.130$ for Lod-lille and $-0.129$ for Jev-1.13. The sign
is not even consistent across arms, and on the commercial arm the hook is
associated with \emph{less} capture, which is the opposite of the predicted
direction.

The sharper test restricts both language cells to the $1{,}429$ items where the
hook is absent and re-reads the gap there (lower block). It barely moves: $+0.358$,
$+0.610$ and $+0.310$, against headline gaps of $+0.352$, $+0.677$ and $+0.256$.
For two of the three arms the gap is slightly \emph{larger} once the hook is
removed. We read this as showing that asymmetry (iv) does not account for the
Bengali--English difference, not as showing the cells are equivalent; a corpus
paired by translation rather than by annotator judgment would settle what we can
only bound here.

\begin{table}[t]
\centering\small
\renewcommand{\arraystretch}{1.12}
\begin{tabularx}{\columnwidth}{@{}>{\raggedright\arraybackslash}X>{\centering\arraybackslash}p{1.05cm}>{\centering\arraybackslash}p{1.05cm}>{\centering\arraybackslash}p{1.05cm}@{}}
\toprule
& \textbf{Laya} & \textbf{Lod} & \textbf{Jev} \\
\midrule
\multicolumn{4}{@{}l}{\emph{Bengali \LAR{} by surface hook}} \\
hook absent ($n=1429$)  & 0.716 & 0.845 & 0.376 \\
hook present ($n=1836$) & 0.742 & 0.975 & 0.247 \\
difference              & $+0.026$ & $+0.130$ & $-0.129$ \\
\addlinespace
\multicolumn{4}{@{}l}{\emph{Capture gap, hook-absent items only}} \\
Bengali \LAR            & 0.716 & 0.845 & 0.376 \\
English \LAR           & 0.358 & 0.236 & 0.066 \\
gap                    & $+0.358$ & $+0.610$ & $+0.310$ \\
\addlinespace
headline gap (all pool) & $+0.352$ & $+0.677$ & $+0.256$ \\
\bottomrule
\end{tabularx}
\caption{\textbf{The content-word hook does not carry the language gap.}
Columns are Laya-ml, Lod-lille and Jev-1.13. ``Hook'' means the Bengali literal
gloss shares at least one content word with the idiom, one of the two
asymmetries most likely to inflate Bengali capture. Upper block: capture is not reliably higher
when the hook is present, and on the commercial arm it is lower. Lower block:
restricting both language cells to hook-absent items leaves the capture gap
essentially unchanged, and larger for two of three arms. The script is released
with the per-item outputs.}
\label{tab:hook}
\end{table}

\paragraph{Testing asymmetry (ii).}
The capture matrix cannot test this, because every item in its pool carries a
parenthetical by construction. The Bengali arm of E1 can. It runs the \Dlit{}
family over all $8{,}876$ items, and on $5{,}485$ of them the literal gloss
carries no parenthetical at all, so the Latin-script cue is simply absent. If the
cue were what produces capture, those items should show little of it.

They show a great deal of it (Table~\ref{tab:paren}). Two things follow, and they
point in opposite directions.

First, the mechanism is real, at least on the marker-based scorer. Laya-ml's
\LAR{} is $0.746$ where the cue is present against $0.581$ where it is absent, a
difference of $+0.165$ with a bootstrap interval of $[+0.145, +0.184]$; on
Lod-lille it is $+0.037$ with an interval straddling zero. Part of the Bengali
capture we report is therefore attributable to a cue rather than to a reading of
the expression, and we would not have found that without looking.

Second, it is not what produces the result. With the cue entirely absent, Laya-ml
still selects the literal gloss on $58.1\%$ of items and Lod-lille on $87.1\%$,
against a chance level of $0.250$ and against English-cell capture of $0.379$ and
$0.241$. The effect is also specific: on the three families that contain no
literal option, the same split moves accuracy by at most $0.042$ with inconsistent
sign, so the parenthetical stratum is not simply the easier or harder half of the
corpus.

\begin{table}[t]
\centering\small
\renewcommand{\arraystretch}{1.12}
\begin{tabularx}{\columnwidth}{@{}>{\raggedright\arraybackslash}X>{\centering\arraybackslash}p{1.6cm}>{\centering\arraybackslash}p{1.6cm}@{}}
\toprule
& \textbf{Laya} & \textbf{Lod} \\
\midrule
\multicolumn{3}{@{}l}{\emph{Bengali \LAR{} under \Dlit}} \\
cue absent    & 0.581 & 0.871 \\
$n$           & 5{,}485 & 560 \\
cue present   & 0.746 & 0.909 \\
$n$           & 3{,}391 & 328 \\
difference    & $+0.165$ & $+0.037$ \\
95\% CI       & $[.145,.184]$ & $[-.005,.077]$ \\
\addlinespace
\multicolumn{3}{@{}l}{\emph{same split, families with no literal option}} \\
\Drand        & $+0.014$ & $-0.028$ \\
\Dsurf        & $+0.025$ & $-0.020$ \\
\Dsem         & $+0.030$ & $+0.042$ \\
\bottomrule
\end{tabularx}
\caption{\textbf{The Latin-script cue is real but does not carry the result.}
\swL Laya-ml, \swD Lod-lille. Items are split by whether the idiom's literal gloss
carries an English parenthetical, which is the only thing in a Bengali option list
written in Latin script. Upper block: the cue is worth $+0.165$ of capture on the
marker scorer and little on the independent-option scorer, yet capture without it
is still $0.581$ and $0.871$ against a chance level of $0.250$. Lower block: the
same split moves accuracy on families containing no literal option by at most
$0.042$, and on Laya-ml by a consistent $1.4$ to $3.0$ points, five to twelve
times less than what it buys on the literal family. The commercial arm cannot be split this way,
since its only Bengali literal run is the pool, where every item carries a
parenthetical.}
\label{tab:paren}
\end{table}

This bounds the cue rather than removing it. It covers the two open arms and not
the commercial one, and the no-cue stratum is a different item set from the pool,
so these numbers are not a clean subtraction from the matrix. The experiment that
would settle it is to strip the parenthetical from the Bengali literal option and
re-run the Bengali cell, which we have not done.

\paragraph{Item construction, and what it could explain.}
\label{app:items}
The literal gloss is an annotated field rather than a string we wrote, and in this
corpus it is often lexically close to the idiom. Of $8{,}876$ items, $3{,}752$
($42.3\%$) have a single-word idiom, $3{,}333$ have two words and the rest three
or more. On $1{,}303$ items ($14.7\%$) the Bengali part of the literal gloss is
identical to the idiom, and on $1{,}599$ ($18.0\%$) it contains the idiom string
verbatim. The literal option averages $41.1$ characters against $31.7$ for the
gold gloss and is the longer of the two on $67\%$ of items, although against all
three alternatives it is the longest on only $40.2\%$. Some entries are also
better described as polysemous words than as idioms: for \bangla{চারপায়া}
(\emph{charpaya}), ``four-legged'' and ``rope cot'' are two senses of one word
rather than a literal and a figurative reading of a phrase. We did not filter
these out, because the corpus's own annotation is what defines the literal and
figurative options and we did not want to substitute our judgment for it, but a
reader should know they are in the pool.

\LAR{} by idiom length is not what a string-matching account predicts. For
Laya-ml it runs $0.635$, $0.684$, $0.595$ and $0.573$ across one-, two-, three-
and four-or-more-word idioms, peaking in the middle rather than on the
single-word items where the idiom and its gloss are most often the same string.
Verbatim overlap does matter: \LAR{} is $0.767$ where the gloss contains the
idiom against $0.616$ where it does not ($+0.151$, bootstrap
$[+0.128, +0.175]$), and $0.986$ against $0.866$ on Lod-lille
($+0.120$, $[+0.089, +0.150]$). Removing both lexical cues at once, which leaves the
$2{,}480$ items that are multiword, carry no verbatim overlap and carry no
English parenthetical, leaves \LAR{} at $0.560$ and $0.891$ and accuracy at
$0.187$ and $0.079$ (Table~\ref{tab:construction}).

Length is not removed by that subset, so we test it on its own. Reconstructing each
\Dlit{} option list under the run's own seed, the literal option is the longest of
the four on $40.2\%$ of all items and averages $41.1$ characters against $31.5$ for
the other three. Inside the cue-free subset both gaps narrow, to $30.6\%$ and
$36.6$ against $32.4$ characters, so the subset does attenuate length without
eliminating it. It does not need to: splitting on whether the literal option is the
longest gives \LAR{} $0.623$ against $0.657$ on Laya-ml, a difference of $-0.034$
$[-0.055, -0.013]$ in the \emph{opposite} direction to the cue, and $0.897$ against
$0.877$ on Lod-lille, $+0.020$ $[-0.024, +0.064]$. Within the cue-free subset the
two differences are $-0.026$ $[-0.069, +0.016]$ and $+0.056$ $[-0.025, +0.130]$,
neither distinguishable from zero. Length is a genuine asymmetry in the material
and not a route to capture, which is why we report it separately rather than
claiming the subset removes it. The worked examples in
Table~\ref{tab:bnexamples} are drawn from this subset so that none of them rests
on verbatim overlap or the Latin-script cue. All five do share a content word with
their literal gloss, which is asymmetry (iv), and Table~\ref{tab:hook} shows that
overlap does not carry capture.

\paragraph{What this appendix does not establish.} Asymmetry (i) remains untested:
we cannot ask what the arms would do on an English state that is a literal
translation rather than an idiomatic equivalent without running that condition. We
flag it in Limitations rather than dismissing it.

\section{Cross-lingual arms}
\label{app:crossling}

\begin{figure*}[!tbp]
\centering
\includegraphics[width=0.74\linewidth]{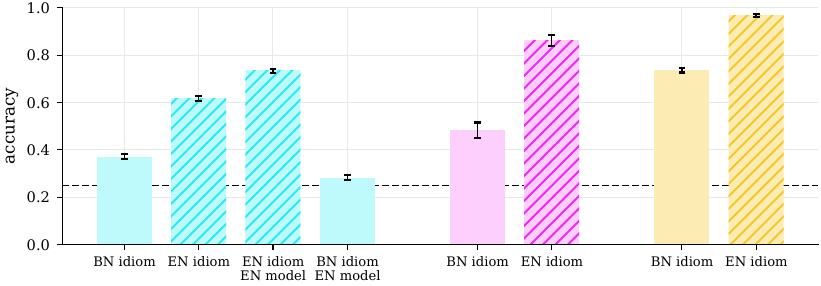}
\caption{Accuracy by task language, grouped by model: \swL Laya-ml (four bars),
\swD Lod-lille, \swJ Jev-1.13. \textbf{Solid} bars are Bengali input,
\textbf{hatched} bars English; Laya-ml's third and fourth
bars use the English-only checkpoint, the fourth being the control. Error bars
95\% bootstrap CIs, dashed line chance. Every model gains between 23 and 38
points when the paired English entry is presented instead of the Bengali idiom.}
\label{fig:crosslingual}
\end{figure*}

Figure~\ref{fig:crosslingual} plots the E4 conditions discussed in
\S\ref{sec:crosslingual}. E4 differs from the capture matrix in two ways that are
easy to conflate, so we state them. E4 uses the \Drand{} family, with no literal
gloss anywhere in the option list, and it asks what the language costs on an
ordinary benchmark item. The matrix uses \Dlit{} and asks what the language does
to literal capture specifically. The two therefore measure different things and
the figure should not be read as a version of Table~\ref{tab:matrix}.

\paragraph{What the four Laya-ml bars are.} Laya-ml has two checkpoints in play.
The first two bars are the multilingual checkpoint on Bengali and on English, which
is the within-model language contrast. The third and fourth use the vendor's
English-only checkpoint, and exist as a control rather than as a result: the third
runs it on the English condition, where it should and does perform, and the fourth
runs it on Bengali script, where it sits at $0.283$, barely above the $0.250$
chance line. That fourth bar does one specific job. A checkpoint documented as
unable to read the script fails here in the documented way, while the multilingual
checkpoint on the same items reaches $0.371$, so the scoring harness does pass
Bengali through to a model that can read it. What the bar cannot do is rule out a
harness that depresses Bengali scores without erasing them, since an English-only
checkpoint would fail on Bengali either way; it shows the harness is not erasing
the language, not that the harness contributes nothing to the deficit.

\paragraph{The gains are not uniform, and the ordering is informative.} Lod-lille
gains the most ($+0.380$), Laya-ml and Jev-1.13 roughly similar amounts ($+0.246$
and $+0.230$). Lod-lille is the arm whose documentation states English-only
training evaluation, and it is also the arm whose Bengali accuracy ($0.484$) is
closest to chance relative to its English accuracy ($0.864$). The natural reading
is that its figurative knowledge is largely intact and the Bengali interface is
where it is lost. We cannot confirm that without probing the model, which we do
not do here, and we note the alternative: a model may also be worse at Bengali
figurative semantics specifically, which would produce the same external
behavior.

\paragraph{Why the error bars matter here.} The three arms run at very different
$n$, so the intervals differ by a factor of three in width. Lod-lille's English
condition in particular runs on its $10\%$ subset. All three gaps clear McNemar at
$p<0.001$ on paired items, which is the test to read rather than the overlap of
the bars, since the bars are unpaired marginals and a paired effect can be
significant while marginal intervals overlap.

\begin{figure}[t]
\centering
\includegraphics[width=\columnwidth]{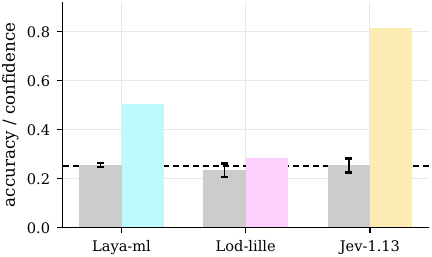}
\caption{The semantics-free control. Per model, the \swG bar is accuracy (95\%
bootstrap CI) and the colored bar is mean reported confidence (\swL Laya-ml,
\swD Lod-lille, \swJ Jev-1.13); the dashed line is the four-option chance level.
No option carries information, so the gap between the two bars is overconfidence:
$0.250$, $0.049$ and $0.561$. Jev-1.13's condition is English.}
\label{fig:nosignal}
\end{figure}

\section{The matched three-way condition}
\label{app:threeway}

\begin{table}[t]
\centering\small
\renewcommand{\arraystretch}{1.12}
\begin{tabularx}{\columnwidth}{@{}>{\raggedright\arraybackslash}X>{\centering\arraybackslash}p{1.05cm}>{\centering\arraybackslash}p{1.25cm}>{\centering\arraybackslash}p{2.05cm}@{}}
\toprule
\textbf{Model} & $n$ & \textbf{Accuracy} & \textbf{95\% CI} \\
\midrule
Laya-ml   & 8{,}876 & 0.367 & [0.357, 0.377] \\
Lod-lille & 888     & 0.484 & [0.453, 0.518] \\
Jev-1.13  & 8{,}761 & 0.736 & [0.727, 0.745] \\
\bottomrule
\end{tabularx}
\caption{\textbf{The common baseline across all three arms.}
The condition is matched across models: a Bengali idiom followed by its
Bengali figurative gloss with random distractors, the same instruction string and
the same $k$. It is not byte-identical, because the commercial arm drew its
sampled distractors under a disjoint seed namespace
(Appendix~\ref{app:repro}), so the three option lists share the gold and differ
in the three distractors. Each row is on that arm's own items, so $n$ differs.
The pairwise differences are computed on the items each pair shares, which is why
they do not equal the differences between rows: Jev--Laya $+0.369$,
Jev--Lod $+0.259$ and Lod--Laya $+0.133$ (McNemar, all $p<10^{-9}$).}
\label{tab:threeway}
\end{table}

\begin{figure}[t]
\centering
\includegraphics[width=\columnwidth]{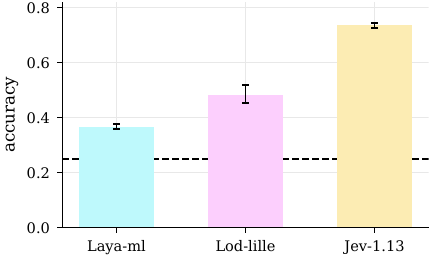}
\caption{The matched three-way comparison. Bars left to right: \swL Laya-ml,
\swD Lod-lille, \swJ Jev-1.13. Black dashed line is chance, error bars 95\%
bootstrap CIs. This capability
ordering is the baseline against which the calibration failures in
\S\ref{sec:confidence}--\S\ref{sec:nosignal} should be read.}
\label{fig:threeway}
\end{figure}

Figure~\ref{fig:threeway} plots Table~\ref{tab:threeway}, the one condition
that is matched across all three arms: Bengali idiom, Bengali glosses, random
distractors, four options, gold position randomized, one instruction string. The
sampled distractors are not the same three, because the commercial arm's blocks
ran under their own seed namespace (Appendix~\ref{app:repro}); what is held is
the construction, not the byte string.

\paragraph{Why only one condition qualifies.} Every other comparison in the paper
is scoped by something the three arms do not share. The literal condition needs
an annotated literal gloss, which exists in Bengali for the whole corpus but in
English only for the $3{,}265$-item pool. The confidence results need a comparable
confidence signal; the maximum probability is the only one all three define
identically (Appendix~\ref{app:payload}), and it is uncalibrated on Laya-ml, whose
shipped per-bucket temperatures were not applied (Appendix~\ref{app:repro}). Beyond
it the arms expose different auxiliary signals, Lod-lille's separate confidence head
and Jev-1.13's presence-of-answer field, which we report where they bear on a
result (\S\ref{sec:confidence}, \S\ref{sec:nosignal}). The order experiment
needs the full permutation set, which
we ran at different $n$ per arm. This condition needs none of that, so it is the
one place where a bare capability ordering can be read without a caveat attached.

\paragraph{What the ordering is for.} It exists to be contradicted later. The
ordering is Jev-1.13 $0.736$, Lod-lille $0.484$, Laya-ml $0.367$, with every
pairwise McNemar significant at $p<10^{-9}$, so it is not a close call. Three
later results do not follow it. Capture is highest on the middle arm
(Table~\ref{tab:matrix}). Calibration is best on the strongest arm by ECE and
worst on the strongest arm on the semantics-free control, where it reports $0.815$
confidence at $0.254$ accuracy. Order stability is highest on the two arms with the
independent-option and commercial designs, which are the middle and the strongest
arm, and lowest on the weakest. Without a baseline ordering none of those
non-correspondences would be statable, which is why a table this simple earns its
own appendix.

\paragraph{The one caveat.} Jev-1.13's cell runs on $8{,}761$ items against
Lod-lille's $888$, and the three arms are not the same size, architecture or
license. The condition is matched on the \emph{task}, not on the models, and a
capability ordering across three systems chosen for architectural contrast is not
a benchmark ranking. We use it as a reference point internal to this paper and
would not quote it as one.

\section{Unit integrity}
\label{app:unit}

\begin{figure*}[!tbp]
\centering
\includegraphics[width=0.80\linewidth]{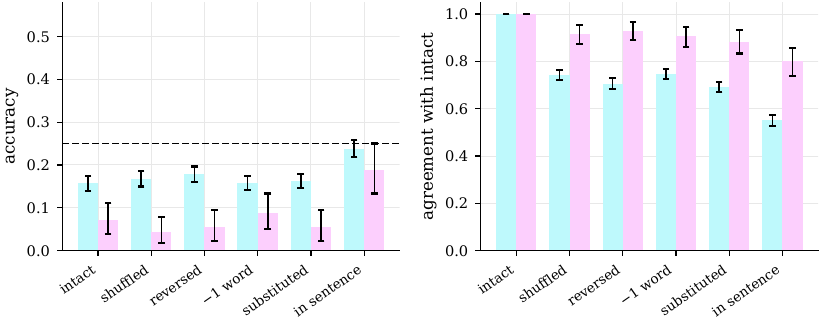}
\caption{Unit integrity: the idiom is perturbed, the options held fixed.
\swL Laya-ml, \swD Lod-lille; error bars 95\% bootstrap
CIs, dashed line chance (left panel). \emph{Left:} accuracy. \emph{Right:}
agreement with the answer on the intact idiom. Scrambling and reversing the
expression does not meaningfully hurt either model, which is what a model reading
the expression as a set of words would show.}
\label{fig:unit}
\end{figure*}

Figure~\ref{fig:unit} plots the E3 perturbations described in \S\ref{sec:bagofwords}.
The options are byte-identical across all six variants, so the only difference
reaching the model is the state string. Table~\ref{tab:bnmanip} shows the six
states for one idiom exactly as they were logged.

\paragraph{Reading Table~\ref{tab:bnmanip}.} The table is the one place in the
paper where the actual strings reaching the model are shown, and it is worth
reading closely because two of its rows look like typesetting errors and are not.
The \emph{word shuffle} and \emph{reversed} rows carry the same string: for a
three-word expression a particular shuffle and the reversal coincide, and the
harness logged both conditions independently and happened to produce the same
state. We print what was logged. Equally, the \emph{substitute one word} row
replaces the final word with an unrelated token, producing a string that is not
Bengali in any idiomatic sense, and the model's probability on the gold gloss
\emph{rises} from $0.21$ to $0.24$. That single cell is the clearest illustration
of \S\ref{sec:bagofwords}: damaging the expression beyond recognition does not cost the
model anything, because the expression was not what it was reading.

The one row that moves is the carrier sentence, and it moves \emph{downward} here
($0.21 \rightarrow 0.07$) even though context raises accuracy on average by
$0.080$ and $0.117$. A single item is not evidence against the aggregate; we
include it rather than hunting for an item that agrees with the mean, because the
aggregate is the claim and the item is an illustration.

The lower block shows the E6 forms applied to the gold option. The scrambled row
is worth comparing word by word with the full gloss: it contains the same lexical
inventory in an order no Bengali reader would accept, and costs the models almost
nothing (Table~\ref{tab:options}). The last row is the semantics-free control, and
the table makes concrete what that control removes, which is everything except the
option's label.

\paragraph{What each perturbation is supposed to break.} The six states form a
graded attack on the expression's integrity. \emph{Intact} is the baseline.
\emph{Word shuffle} destroys word order while preserving the lexical inventory,
so a model storing the expression as an ordered unit should fail while a model
matching its words should not. \emph{Reversed} is the specific shuffle that is
maximally far from the original in order while being deterministic rather than
sampled. \emph{Delete one word} removes material, which should hurt any model
relying on the full form. \emph{Substitute one word} replaces material with an
unrelated token, producing a string that is not the idiom at all and is not
idiomatic Bengali. \emph{Carrier sentence} adds the corpus's own example sentence
around the intact idiom, which adds disambiguating context without touching the
expression.

\paragraph{Reading the two panels together.} Accuracy on the left and
intact-agreement on the right answer different questions, and either alone is
misleading. Accuracy can stay flat because the perturbation does nothing, or
because it changes many answers in compensating directions. Agreement
distinguishes those: it counts the share of items whose answer is unchanged from
the intact state, so a model that is unaffected scores high on both while a model
that is churning scores flat accuracy and low agreement. Laya-ml agrees with
itself on $0.742$ of items under shuffling and Lod-lille on $0.917$. For Lod-lille
the perturbation is close to invisible. For Laya-ml it is not negligible, but it is
no larger than what a meaning-preserving rewrite does: the annotators' certified
alternative forms leave Laya-ml's answer unchanged on only $0.657$ of pairs
(Appendix~\ref{app:surface}). Scrambling the expression's word order moves this
model no more than paraphrasing it faithfully, which is the opposite of what a
model holding the expression as an ordered unit would show.

\paragraph{What this rules out, and what it does not.} It rules out the
hypothesis that these models hold the idiom as a stored lexical unit whose
retrieval is keyed on its exact form, because breaking the form does not cost them
the answer. It does not establish what they do instead. The state string is
read: \S\ref{sec:construction} shows a literal gloss that contains the idiom
verbatim is chosen more often. What the shuffle shows is that its word order
carries little, which fits a model that reads the expression as a set of words. The
carrier sentence is the only manipulation that moves either arm, by $+0.080$ and
$+0.117$, and it is the one that adds context rather than altering the expression.
Our reading is that the usable figurative signal is in the surrounding material
rather than in the expression's form, and we state it as a reading.

\paragraph{What E3 can and cannot show.} Three scope conditions bound it, and we
state them rather than leaving them to be inferred.

\emph{The perturbations are real.} E3 selects idioms of three words or more under
the same tokenizer the distractor builder uses. The corpus holds $1{,}791$ such
idioms and the logged outputs cover $1{,}778$ of them, $1{,}136$ of three words,
$495$ of four and $147$ longer; the $13$ missing are absent from the logged output
rather than excluded by the rule, and at $0.7\%$ of the condition they move no
reported figure. No manipulation is a no-op on a one- or two-word string. Measured against the logged intact state, the
shuffle changes the string on $99.9\%$ of items, reversal on $99.7\%$, deletion and
substitution on $100\%$. The one genuine collision is between shuffle and reversal,
which coincide on $16.0\%$ of items because a three-word reversal is one of only
six permutations; we report both rows rather than merging them.

\emph{Accuracy is floored.} The condition is \Dlit{}, where intact accuracy is
already $0.157$ and $0.072$, below the $0.250$ chance level. A perturbation that
destroyed the figurative reading entirely could not push accuracy much lower, so
the flat accuracy rows are weak evidence on their own and we do not rest the
conclusion on them. Agreement with the intact answer has no such floor: it can run
from $0$ to $1$ regardless of how accurate either answer is, and it is $0.742$ and
$0.917$ under shuffling. That is the number the ``not a unit'' reading rests on.
Running E3 under \Drand, where there is headroom above chance, would let accuracy
carry the argument too, and we did not.

\emph{Power.} Lod-lille's E3 runs on $180$ items, so several of its individual
perturbation contrasts are not separately significant and we do not claim them.
The claim is the pattern across six conditions and two arms, and the one contrast
that is individually large in both.

\section{Surface invariance}
\label{app:surface}

The corpus ships alternative surface forms certified by annotators as
meaning-preserving: a variant spelling, a clipped form, or a near-synonymous
rewording recorded in the \texttt{alternative\_idioms} field by the same experts
who wrote the glosses. These are not perturbations of ours. A model reading meaning
rather than form should return the same answer on both members of such a pair. E2
presents each pair under the \Dhard{} family, so the literal gloss is in the option
list for both members and the comparison is made under the condition where capture
is strongest.

Laya-ml agrees with itself on $0.657$ of $3{,}143$ pairs ($2{,}423$ items) and
Lod-lille on $0.677$ of $310$ pairs ($234$ items): about a third of the time, a
rewrite the annotators certified as meaning-preserving changes the answer. Mean
Jensen--Shannon divergence between the two returned probability vectors is $0.039$
and $0.085$, on a scale where $0$ is identical and $1$ is disjoint support.

\paragraph{Reading agreement against divergence.} The two numbers look
inconsistent and are not. Small mean divergence with frequent answer changes is the
signature of decisions taken on thin margins: the rewrite moves the distribution
only slightly, but enough to cross the argmax boundary. Laya-ml's mean peak
probability on these pairs is $0.535$, so its top two options are often close and a
small shift reorders them (Appendix~\ref{app:dists} gives the full distributions). Lod-lille is the harder case at $0.737$: its
distributions are sharp, yet it still changes answer on a third of pairs, so for
that arm the thin-margin reading covers part of the instability rather than all of
it.

\paragraph{Why this is reported as a control rather than a headline.} E2 is the
experiment designed to explain the E1 result away, and it does not. If capture were
an artifact of the particular surface form the corpus happens to record, presenting
the annotators' alternative form should recover the figurative reading. It does not.
Accuracy does move, and significantly: $0.138 \rightarrow 0.193$ for Laya-ml and
$0.035 \rightarrow 0.158$ for Lod-lille (McNemar exact $p < 0.001$ on both), so the
surface form is not irrelevant. These canonical-form accuracies are lower than the
\Dhard{} rows of Table~\ref{tab:fullladder} ($0.144$ and $0.063$) because E2 runs
only on the items that carry an annotated alternative form, $2{,}423$ of $8{,}876$
and $234$ of $888$, and that subset is harder for both arms; the comparison here is
within the pair, so the level does not affect it. But both forms sit far below the $0.250$ chance
level on both arms, so the alternative form rescues neither model. We read the two
results together: these scorers are genuinely unstable under meaning-preserving
rewriting, which is worth reporting on its own, and that instability is neither
large enough nor in the right direction to be the mechanism behind a \LAR{} of
$0.644$ to $0.885$.

\paragraph{Caveat on the pairs.} Items vary in how many alternative forms they
carry, so the pair counts exceed the item counts and items with more variants are
weighted more heavily. We report the pair-level figure because it is the
quantity a caller experiences; averaging within item first gives $0.667$ and
$0.697$, so the pair weighting \emph{lowers} agreement by $1.0$ and $1.9$ points,
which is why the pair-level figure is the more conservative of the two.

\section{Subset representativeness}
\label{app:subset}

Lod-lille ran a fixed $10\%$ random subset, drawn once with seed $909$ and reused
for every experiment on that arm except the capture matrix, which it ran on the
full $3{,}265$-item pool, so its subset cells are comparable to each other.
The question this appendix answers is whether they are comparable to the other
arm's full-set cells. Because the subset is a subset of the same corpus, we can
measure the answer directly rather than assume it: Table~\ref{tab:subset} runs
Laya-ml, which has the full set, on those same $888$ items and compares it with
itself.

\paragraph{Why this is the right test.} An alternative would be to compare
Lod-lille on the subset against Laya-ml on the full set and attribute any
difference to sampling, which conflates the sample with the model. Holding the
model fixed and varying only the sample isolates the sampling term. The drift in
Table~\ref{tab:subset} measures how well the subset stands in for the full corpus.
It does not enter the paired cross-arm comparisons, which are made on the shared
888 items where both arms see the same subset.

\paragraph{What the drift column shows.} The largest absolute drift across five
families is $0.016$, on \Drand{}. Three families drift down and two up, but the
two upward drifts are negligible ($+0.001$, $+0.004$), so the fair summary is a
slight downward lean of under two points rather than no lean at all. Two of the five subset estimates
fall marginally outside the full-set bootstrap interval, which at $n=888$ against
$n=8{,}876$ is the expected rate rather than a signal. Critically, no family's
drift is large enough to reorder the ladder: \Dlit{} and \Dhard{} remain far below
the other three on both the full set and the subset, and the \Drand{} to \Dlit{}
gap, which is the quantity the paper reports, changes by $0.001$.

\begin{table}[t]
\centering\small
\renewcommand{\arraystretch}{1.12}
\begin{tabularx}{\columnwidth}{@{}>{\raggedright\arraybackslash}X>{\centering\arraybackslash}p{1.35cm}>{\centering\arraybackslash}p{1.35cm}>{\centering\arraybackslash}p{1.15cm}@{}}
\toprule
\textbf{Family} & \textbf{Full} & \textbf{Subset} & \textbf{Drift} \\
\midrule
\Drand & 0.367 & 0.351 & $-0.016$ \\
\Dsurf & 0.338 & 0.339 & $+0.001$ \\
\Dsem  & 0.333 & 0.325 & $-0.007$ \\
\Dlit  & 0.151 & 0.136 & $-0.015$ \\
\Dhard & 0.144 & 0.149 & $+0.004$ \\
\bottomrule
\end{tabularx}
\caption{Laya-ml on the 10\% subset against its own full set, drift computed
before rounding. Maximum drift is $0.016$, and the larger drifts are downward,
a slight lean of under two points. Two of five subset estimates fall marginally outside the full-set
bootstrap CI, which is expected at $n=888$ and changes no sign or ordering.
The full-set figures are therefore reported as the precise estimates, and
Lod-lille's intervals are reported wider throughout.}
\label{tab:subset}
\end{table}

\section{Full option ablation}
\label{app:options}

\begin{figure*}[t]
\centering
\includegraphics[width=0.90\linewidth]{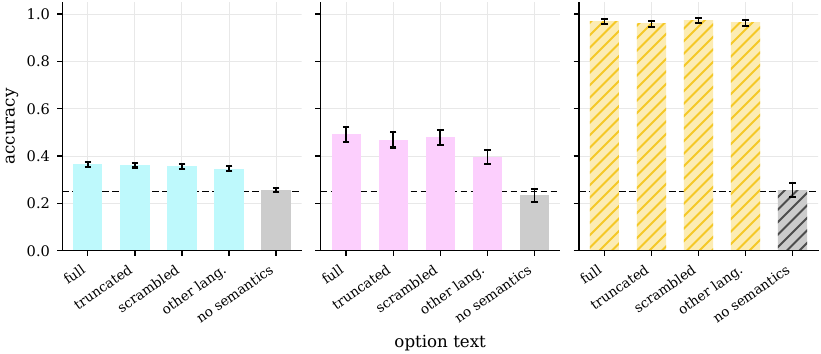}
\caption{Option-side ablation. Panels left to right: \swL Laya-ml, \swD Lod-lille,
\swJ Jev-1.13. In each panel the fifth, \swG bar is the semantics-free control
of \S\ref{sec:nosignal}; the whole Jev-1.13 panel is
hatched because its task is in \textbf{English}. Error bars are 95\% bootstrap
CIs and the black dashed line is chance. Truncating and word-scrambling the
option text cost almost nothing in any of the three models.}
\label{fig:options}
\end{figure*}

Table~\ref{tab:options} gives the numbers behind Figure~\ref{fig:options}. The
five rows hold the state, the instruction and the gold position fixed and rewrite
every option in the list the same way, so a row is a statement about what the
option text needs to contain for the decision to survive.

\paragraph{The first three rows are the finding.} Truncating each option to its
first six words costs $0.004$, $0.024$ and $0.011$ across the three arms.
Scrambling the words \emph{inside} each option, which leaves a string no reader of
either language would accept as a sentence, costs $0.008$, $0.013$ and nothing at
all; on Jev-1.13 the scrambled row is marginally \emph{above} the full gloss, which
is within noise but makes the point that the manipulation is invisible to it. Taken
together: what these models need from an option is its lexical inventory, and the
order and syntax of that inventory are close to free. This is the behavioral half
of the result \citet{sun2026typesafe} reach from the other direction by permuting
which option name is bound to which definition.

\paragraph{The fourth row is a separate effect.} Presenting the options in the
other language, with the state unchanged, costs Laya-ml $0.018$ and Jev-1.13
$0.007$ but Lod-lille $0.095$, which is five to thirteen times as much. The arm
that loses most from a language mismatch between state and options is the arm whose
documentation reports English-only training evaluation, and the one with the
largest cross-lingual gap in E4. We note the correspondence; with three models we
cannot test it.

\paragraph{The fifth row is the control, and it is the row that validates the
instrument.} Replacing every option with a bare label leaves $0.256$, $0.233$ and
$0.254$, against a chance level of $0.250$. This matters more than it looks. It
shows that the four preceding rows are measuring option semantics rather than some
artifact of the harness, because when the semantics are removed the measurement
collapses to exactly where it should. It also shows the gold-position randomization
is working: a harness with a position leak would score above chance here. And it is
the condition \S\ref{sec:nosignal} builds on, because it is the one place where
accuracy is known a priori, at $0.25$, so any reported confidence above that is
unambiguously excess.

\paragraph{Caveats.} Three bound what the \textsc{o}-series shows. First,
truncation to six words is a weaker manipulation than it sounds: the median Bengali
gloss is four words, so truncation leaves $77.8\%$ of options untouched, and the
English options, median seven words, are cut on $58.0\%$. The scramble is the
informative row: it is necessarily a no-op on the $5.7\%$ of options that are a
single word, and otherwise changes the option's word order. Second, the whole series runs on random distractors, where
identifying the right words already suffices to answer; a model that composes
meaning and one that only matches words would both score well. So these rows show
that composition is not \emph{required} on this condition, not that these models
cannot compose. Running the scramble under \Dlit{} or \Dsem, where the distractors
are lexically close and composition would pay, is the test that would separate the
two, and we did not run it. Third, Jev-1.13's column is English throughout, so its
rows are not comparable in level to the other two, only in shape. The three arms' \emph{full
gloss} rows differ slightly from the corresponding \Drand{} rows of
Table~\ref{tab:fullladder} because E6 draws its own distractor sample, under seed
$6000$ on the open arms and $16000$ on the commercial arm, whose E6 also runs on
$873$ items rather than $3{,}265$; the differences are $0.002$ to $0.008$ and we
report both rather than reconciling them silently.

\begin{table}[t]
\centering\small
\renewcommand{\arraystretch}{1.12}
\begin{tabularx}{\columnwidth}{@{}>{\raggedright\arraybackslash}X>{\centering\arraybackslash}p{1.05cm}>{\centering\arraybackslash}p{1.05cm}>{\centering\arraybackslash}p{1.05cm}@{}}
\toprule
\textbf{Option text} & \textbf{Laya} & \textbf{Lod} & \textbf{Jev} \\
\midrule
full gloss   & 0.365 & 0.492 & 0.970 \\
truncated    & 0.361 & 0.468 & 0.958 \\
scrambled    & 0.357 & 0.479 & 0.974 \\
other lang.  & 0.347 & 0.398 & 0.963 \\
no semantics & 0.256 & 0.233 & 0.254 \\
\bottomrule
\end{tabularx}
\caption{\textbf{Option-side ablation.}
The semantics-free row removes meaningful option content and yields
near-chance accuracy in all three arms. Jev-1.13's column is English.}
\label{tab:options}
\end{table}

\section{Order sensitivity and position bias}
\label{app:position}

This appendix reports the per-slot selection shares behind \S\ref{sec:order},
then sets out the permutation audit that establishes the residual instability of
the independent-option scorer is real rather than an artifact of our harness.

\paragraph{What is held fixed.} Each item here is one idiom with one fixed set of
four option texts. We present that same set 24 times, once per permutation, and
change nothing else: the instruction string, the state, $k$, the decoding
configuration and the option texts themselves are byte-identical across the 24
calls, and no call is retried. The only quantity that varies is which slot, and
therefore which letter, each option occupies. Anything that moves across the 24
calls is attributable to that reordering, which cannot separate position from
labeling (\S\ref{sec:order}): the condition is clean in what it holds fixed, not in
what it can attribute.

\paragraph{Item samples.} The three arms' item samples were drawn separately, $992$
for Laya-ml, $299$ for Lod-lille and $300$ for Jev-1.13, and barely overlap: no
item is common to all three, and any two share at most $33$. Every cross-arm
figure in Tables~\ref{tab:order} and~\ref{tab:orderaudit} is therefore a comparison
of rates on different items, not a paired one. At these sizes the gap between
$83.7\%$ and $10.7\%$ is far outside sampling error, but item composition is not
controlled across arms.

\paragraph{Recovering option identity.} The harness records, per call, the
permutation as a four-digit string in which position $s$ holds the index, in the
canonical option list, of the option displayed in slot $s$. A chosen slot must be
mapped back through that permutation before it can be compared across calls.
Counting distinct slot \emph{letters} instead inflates instability to near unity
for any scorer, because a letter necessarily changes whenever the same gloss is
moved; on Laya-ml that metric reads $1.000$ against the $0.837$ reported here. It
is an easy substitution to make and it is not what we want to measure, so we flag
it. All numbers in the paper use the mapped gloss identity.

\begin{table}[t]
\centering\small
\renewcommand{\arraystretch}{1.12}
\begin{tabularx}{\columnwidth}{@{}>{\raggedright\arraybackslash}X>{\centering\arraybackslash}p{1.05cm}>{\centering\arraybackslash}p{1.05cm}>{\centering\arraybackslash}p{1.05cm}@{}}
\toprule
& \textbf{Laya} & \textbf{Lod} & \textbf{Jev} \\
\midrule
answer changes      & \textbf{0.837} & 0.107 & 0.133 \\
correctness varies  & \textbf{0.686} & 0.070 & 0.120 \\
distinct glosses    & \textbf{2.83}  & 1.14  & 1.15  \\
perfectly stable    & 0.163 & \textbf{0.893} & 0.867 \\
$p$(gold) range     & \textbf{0.346} & 0.029 & 0.128 \\
selection spread    & \textbf{0.139} & 0.030 & 0.011 \\
\bottomrule
\end{tabularx}
\caption{\textbf{Option-order sensitivity.} Each item is evaluated under all 24
orderings of the same four options ($n=992/299/300$).
\swL Laya, \swD Lod, \swJ Jev; bold marks the largest value in each row.
\emph{$p$(gold) range} is the within-item range of the gold option's probability
across the 24 orderings, averaged over items; \emph{selection spread} is the gap
between the most and least chosen slot, pooled over orderings, and is $0$ under no
positional preference. The marker-based scorer changes its chosen gloss on
$83.7\%$ of items against $10.7\%$ for the independent-option scorer. The three
item samples are drawn separately per arm, so cross-arm comparisons are unpaired.
Lod is documented as order-independent by construction; the permutation audit below
shows it is not, which is why its entries here are small rather than zero.}
\label{tab:order}
\end{table}

\paragraph{The permutation audit.} Table~\ref{tab:orderaudit} reports a quantity
that the summary measures in Table~\ref{tab:order} cannot: not whether the
\emph{answer} changes, but how far a \emph{single option's own score} travels as
that option is moved around the list. This is the quantity the independent-option
design makes a claim about. Scoring every option from the same start position, so
that no option attends to any other, should make an option's score a function of
that option and the state alone, so the spread should sit at the level of
floating-point accumulation noise, which for
\texttt{fp32} summation over sequences of this length is on the order of
$10^{-7}$. It does not. Lod-lille's median item has an option whose score moves
$4.0 \times 10^{-2}$, five orders of magnitude higher, and not one of its 299
items falls below $10^{-4}$. Two consequences follow. First, the $10.7\%$
answer-change rate in Table~\ref{tab:order} is not a tie-breaking or
slot-indexing artifact of our harness: ties would require near-identical scores,
and the scores are not near-identical, so what moves is the scoring itself.
Second, the documented guarantee is not one an evaluation can rely on, at least in
the configuration we were able to run. We take the narrow reading throughout: the
independent-option design is a large improvement over the marker-based one on this
axis, and it is not the categorical property its documentation describes.

\begin{table}[t]
\centering\small
\renewcommand{\arraystretch}{1.12}
\begin{tabularx}{\columnwidth}{@{}>{\raggedright\arraybackslash}X>{\centering\arraybackslash}p{1.05cm}>{\centering\arraybackslash}p{1.05cm}>{\centering\arraybackslash}p{1.05cm}@{}}
\toprule
& \textbf{Laya} & \textbf{Lod} & \textbf{Jev} \\
\midrule
items                       & 992   & 299   & 300   \\
median spread               & \textbf{0.495} & 0.040 & 0.070 \\
mean spread                 & \textbf{0.507} & 0.046 & 0.139 \\
95th pct.\ spread           & \textbf{0.772} & 0.105 & 0.480 \\
max spread                  & \textbf{0.928} & 0.245 & 0.820 \\
share $< 10^{-4}$           & 0.000 & 0.000 & \textbf{0.230} \\
answer changes              & \textbf{0.837} & 0.107 & 0.133 \\
\bottomrule
\end{tabularx}
\caption{\textbf{Permutation audit.} Per item, the largest distance a single
option's own score travels across the 24 orderings of its list, after mapping
each slot back to the option it displayed.
\swL Laya-ml, \swD Lod-lille, \swJ Jev-1.13. Bold marks the largest value in each
row. An order-independent scorer should sit at \texttt{fp32} noise, around
$10^{-7}$; the \emph{share} row gives the fraction of items at or below
$10^{-4}$, a threshold three orders of magnitude more permissive than that. Only Jev-1.13 has any items at
that threshold, and then fewer than a quarter; its probabilities are reported to
two decimal places, so for that arm the row means identical \emph{as reported}
rather than identical in the underlying computation. The script that produces this table is released with the
per-item outputs.}
\label{tab:orderaudit}
\end{table}

\paragraph{The commercial arm.} Jev-1.13 makes no public invariance claim, and its
behavior is the least like either open arm. It returns identical distributions on
$23.0\%$ of items, with the caveat above that its probabilities reach us rounded to
two decimal places, so this describes the reported values rather than the
computation behind them. Against that, its mean spread ($0.139$) is three times
Lod-lille's and its maximum ($0.820$) approaches Laya-ml's. The distribution is bimodal rather than merely noisier,
which is what one would expect from a scorer that is close to deterministic on
easy items and switches regime on hard ones. We cannot test that reading without
weights, and we do not claim it.

\begin{figure*}[t]
\centering
\includegraphics[width=0.90\linewidth]{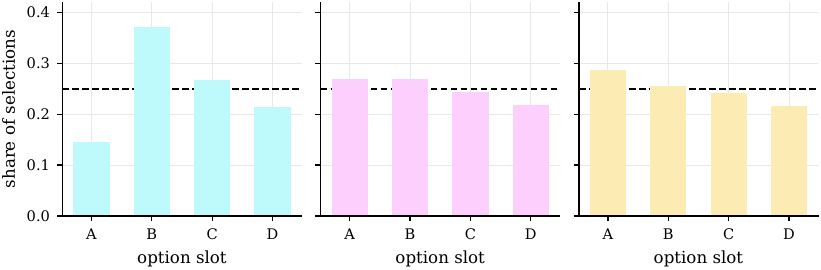}
\caption{Share of selections falling on each option slot, over the full ladder
condition rather than the permutation sweep. Panels left to right:
\swL Laya-ml, \swD Lod-lille, \swJ Jev-1.13. The gold option is placed
uniformly at random, so the black dashed line at $0.25$ is the expectation under
no positional bias and any departure from it is bias. Laya-ml peaks on slot B,
taking it $37.2\%$ of the time against $14.6\%$ for slot A; the other two arms sit
close to the line throughout, with a mild and monotone preference for earlier
slots.}
\label{fig:position}
\end{figure*}

\begin{figure*}[t]
\centering
\includegraphics[width=0.90\linewidth]{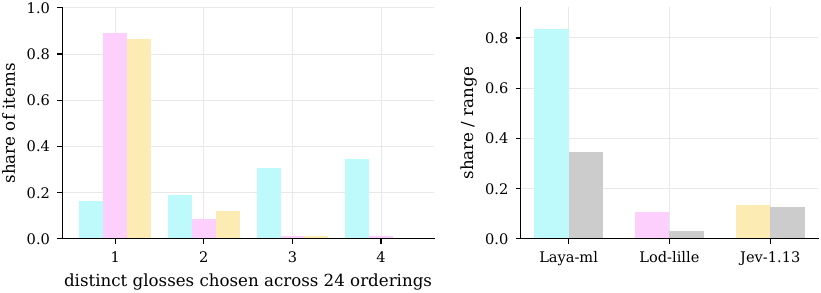}
\caption{Order sensitivity. \emph{Left:} how many distinct glosses each model
selects across the 24 orderings of one option set; within each group bars run \swL Laya-ml,
\swD Lod-lille, \swJ Jev-1.13, and a value of 1 means perfect stability.
\emph{Right:} per model, the colored bar is the share of items whose chosen
gloss is unstable and the \swG bar is the mean within-item range of $p$(gold). Laya-ml is
unstable on 83.7\% of items with a mean $p$(gold) swing of $0.346$; the other two
are near-stable.}
\label{fig:order}
\end{figure*}

\paragraph{Position bias.} Figure~\ref{fig:order} gives the distributional
summaries behind Table~\ref{tab:order}, both from the permutation sweep.
Figure~\ref{fig:position} is the one panel in this appendix drawn from a different
run: it reports per-slot selection shares over the full ladder condition, pooled
over its families, on $8{,}876$ items for Laya-ml, $888$ for Lod-lille and
$8{,}761$ for Jev-1.13, because position is randomized per item there and the
estimate is far better powered than the sweep's $992$ items would allow. The sweep agrees in direction and is milder, $32.7\%$
against $18.8\%$ on slot B and slot A for Laya-ml. Because the gold option is
placed uniformly at random in both, the expectation under no positional bias is
exactly $0.25$ in every slot for every arm, so neither needs a baseline model: the
distance from the dashed line \emph{is} the bias. Laya-ml's profile is not the
primacy gradient usually reported for prompted generative scorers
\citep{zheng2024robust}; it peaks on slot B at $37.2\%$ against $14.6\%$ for slot
A, a $2.5\times$ swing in selection share from position alone, and is not a
monotone function of position at all. We note the shape and
decline to explain it; with one model per scoring design, we have no way to
separate a property of the marker scheme from a property of this particular
backbone.

\section{Run record}
\label{app:repro}

\paragraph{Environment.} Python 3.13.15 throughout. The two open arms ran on one
Tesla T4 (15{,}360\,MiB): Laya-ml with \texttt{amp\_dtype} set to \texttt{bf16}
and \texttt{max\_len} 1024 at 27.2\,ms per call, Lod-lille in \texttt{fp32} at
86.1\,ms per call. The T4 is Turing, whose tensor cores take \texttt{fp16} and
integer inputs only, so there is no native \texttt{bf16} path and no
\texttt{fp32} or \texttt{tf32} tensor-core path either: the autocast context
accepts the setting and the arithmetic falls back to ordinary \texttt{fp32} CUDA
cores, with \texttt{bf16} rounding only at the cast boundaries. We report the configured value because that is what the run record
holds, and flag it because a reader comparing against an Ampere or later card is
not comparing the same numerics.

The commercial arm needs no accelerator and ran on CPU against
\texttt{typesafe/jev-1.13}. That alias is mutable: at the time of our run it
resolved to the \texttt{20260917} build, and a later caller using the same alias
may not reach the same weights. Every commercial figure in this paper should be
read against that build.

\paragraph{Release.} The instrument, the per-item outputs for every cell, the
analysis scripts and the figure sources are available under CC BY-SA 4.0 at
\url{https://github.com/sakhadib/literal-lock-in}, with the project page at
\url{https://sakhadib.github.io/literal-lock-in}.

\paragraph{Temperatures.} Laya-ml's run recorded
\texttt{temperature = [1.0, 1.0, 1.0]} with \texttt{temperature\_by\_options}
empty, so the per-bucket temperatures it ships were not applied and its ECE
figures are raw. Lod-lille ships \texttt{temperature: 1.0} by design, its
authors having reported that fitting hurt held-out calibration. Both open arms
are therefore untempered and comparable on that axis. The commercial arm's
internal calibration is not observable.

\paragraph{Seeds.} The Bengali arms use a per-experiment seed (E1 1000, E2 2000,
E3 3000, E4 4000, E6 6000, E7 7000) added to the item id, so each item's option
permutation is reproducible and identical across arms. The split seed is
20261001. The commercial arm uses a disjoint block seed set (order audit 17000,
literal 12000, ceiling 11000, Bengali control 14000, options 16000) because its
blocks ran in a different order; the permutation for a given item is still a
pure function of seed and id. Lod-lille's 10\% subset was drawn once with seed
$909$.

\paragraph{Call volume and cost.} The commercial arm issued 56{,}533 calls for
\$0.9348 at \$0.042 per million input tokens with output free. The blocks are
the order audit ($300$ items $\times\,24$ permutations $= 7{,}200$), literal
capture on the shared pool ($3{,}265 \times 3$ families, \Drand, \Dlit{} and
\Dhard, $= 9{,}795$), the English ceiling ($8{,}761 \times 3$ families, \Drand,
\Dsurf{} and \Dsem, $= 26{,}283$), the Bengali control ($8{,}761$, one family) and
the option ablation ($873 \times 5$ option forms $= 4{,}365$), which is
56{,}404 retained rows. The option ablation runs on $873$ items rather than the
$888$ of the shared subset because this block is English and $15$ of the $888$
carry no English equivalent idiom to use as the state; the $873$ are exactly the
intersection of the subset with the $8{,}761$ items that do. The remaining 129 calls
are the endpoint probe and the response-contract checks made before each block.
Transient 429 and 5xx responses were retried with exponential backoff to a
30\,s ceiling; no call was dropped.

\paragraph{Statistics.} All intervals are 95\% percentile bootstrap over items,
2{,}000 resamples, fixed seed 0. ECE uses 15 equal-width bins. Paired
comparisons use McNemar's exact test via the binomial distribution. AURC is
reported against a random-gate baseline computed on the same items. Differences
and ratios are computed from unrounded values, so they can differ in the last digit
from differences of the printed figures.

\paragraph{What is released.} The per-item outputs for every cell of every
experiment (one row per item per condition, carrying the full probability
vector, the predicted key, the gold key, the literal key and the gold position),
the five split manifests, the three run environment records quoted above, the
task configurations containing the verbatim instruction strings, the analysis
scripts that turn those outputs into every table in this paper, the figure
sources, and, beside each figure, a CSV of the exact plotted values.

\paragraph{Pairing.} Because the two open arms were reconstructed independently
from a written contract rather than from shared code, the released outputs admit
a direct check: on the 4{,}440 overlapping rows the gold key, gold position,
idiom string and literal-distractor key agree at $1.000$. That check is itself a
released artifact, and it is what licenses every paired test in the paper.

\paragraph{The matrix run.} The six cells of \S\ref{sec:capture} were run as one
job, with their own seed namespace (21000) so no option permutation collides with
the runs above. Every cell covers the same 3{,}265
items and is built by a single function; the two languages share one distractor
draw per item and are shuffled independently from the same generator. Laya-ml
ran both languages on the multilingual checkpoint, so only the language changes
and the checkpoint does not. Lod-lille ran the full pool here rather than its
10\% subset, so its matrix cells rest on $3{,}265$ items. The
commercial arm issued 6{,}531 calls for \$0.1152. Its English cell reproduces
its English-ladder value to four decimal places ($0.0472$), which is the check
that the two constructions measure the same thing.

\paragraph{Known gaps.} E2 was not run on the commercial arm, a gap in coverage
rather than a constraint of language or cost. Cost was not the constraint: the whole commercial arm came to
\$1.05, the \$0.9348 of the main blocks plus the \$0.1152 of the matrix run. The matrix covers the literal condition only; the other distractor
families were not run in both languages. The confidence \emph{inversion} remains a
two-arm result, although \S\ref{sec:confidence} reports the commercial arm's
Bengali calibration from the matrix outputs.

\end{document}